\documentclass[final]{cvpr}

\usepackage{times}
\usepackage{epsfig}
\usepackage{graphicx}
\usepackage{amsmath}
\usepackage{amssymb}

\usepackage[pagebackref=true,breaklinks=true,colorlinks,bookmarks=false]{hyperref}
\usepackage{cleveref}
\usepackage{comment}
\usepackage{capt-of}
\usepackage{booktabs}
\usepackage{multirow}
\usepackage{verbatim}

\usepackage{grffile}

\usepackage{amsthm}
\theoremstyle{definition}
\newtheorem{definition}{Definition}[section]

\newcommand\sizeImage{0.14}

\newcommand\sizeHorizontalSpace{-7pt}

\def\cvprPaperID{****} 
\def\confYear{CVPR 2021}

\begin{document}

\title{DeepBlur: A Simple and Effective Method for Natural Image Obfuscation}

\author{
    Tao Li \\
    Department of Computer Science \\
    Purdue University \\
    {\tt\small taoli@purdue.edu}

    \and

    Min Soo Choi \\
    School of Industrial Engineering \\
    Purdue University \\
    {\tt\small choi502@purdue.edu}
}

\twocolumn[{
\renewcommand\twocolumn[1][]{#1}
\vspace{-6.5pt}
\maketitle
\begin{center}
\setlength{\tabcolsep}{.15em}
\begin{tabular}{ccccccc}
    $Original$ & $Blurring$ & $Pixelation$ & $Masking$ & $Adv Noise$ & $Ours^\dagger$ & $Ours^\ddagger$ \\
    \includegraphics[width=\sizeImage\textwidth]{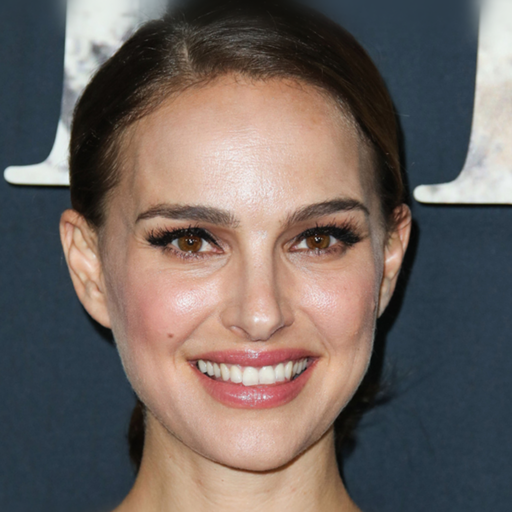}
    & 
    \includegraphics[width=\sizeImage\textwidth]{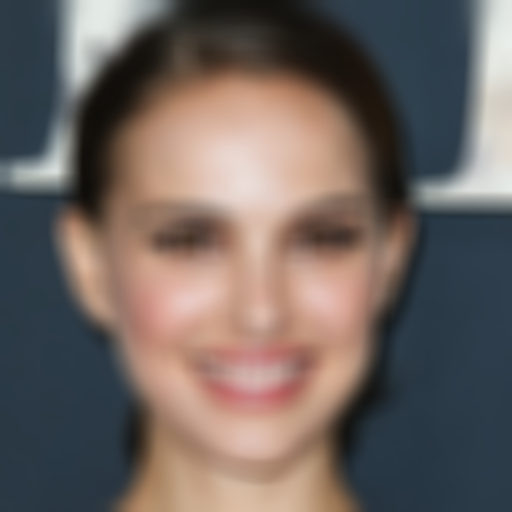}
    & \hspace{\sizeHorizontalSpace}
    \includegraphics[width=\sizeImage\textwidth]{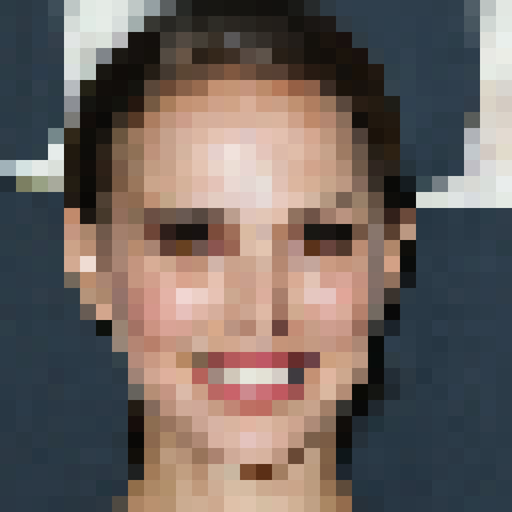}
    & \hspace{\sizeHorizontalSpace}
    \includegraphics[width=\sizeImage\textwidth]{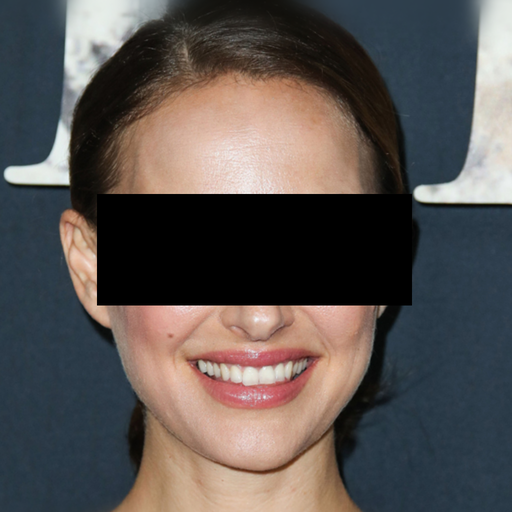}
    & \hspace{\sizeHorizontalSpace}
    \includegraphics[width=\sizeImage\textwidth]{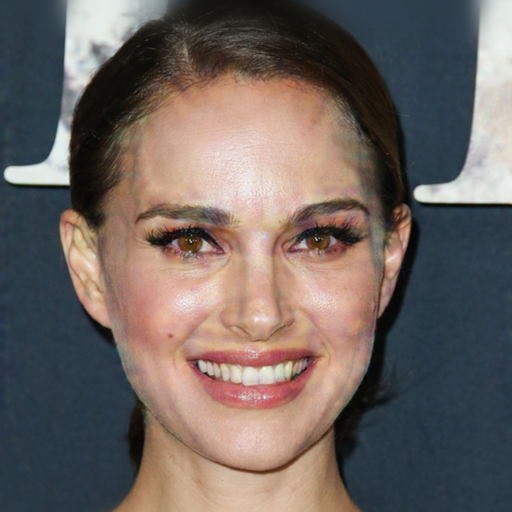}
    & \hspace{\sizeHorizontalSpace}
    \includegraphics[width=\sizeImage\textwidth]{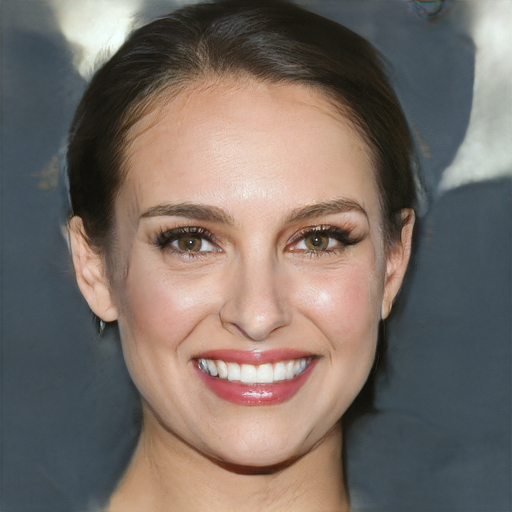}
    & \hspace{\sizeHorizontalSpace}
    \includegraphics[width=\sizeImage\textwidth]{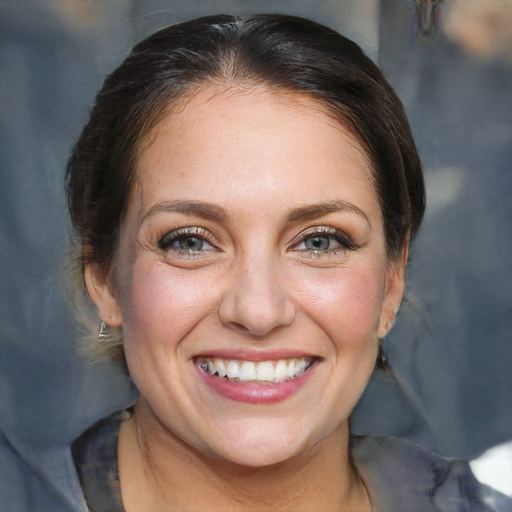}
    \\
    \includegraphics[width=\sizeImage\textwidth]{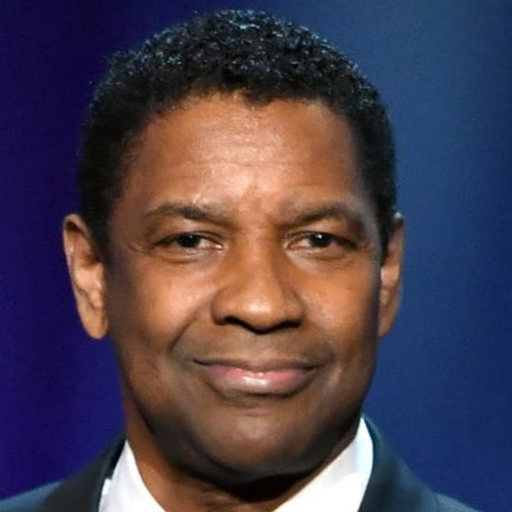}
    & 
    \includegraphics[width=\sizeImage\textwidth]{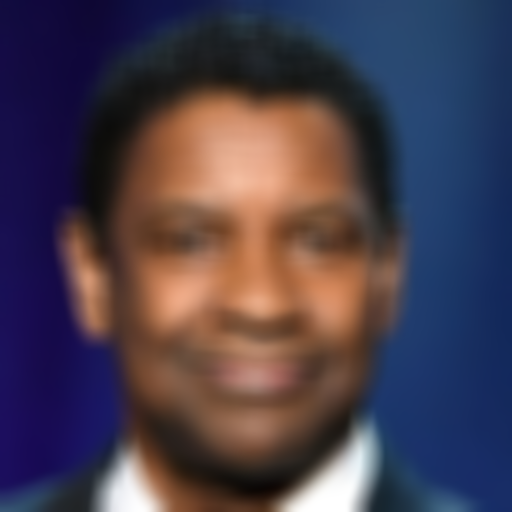}
    & \hspace{\sizeHorizontalSpace}
    \includegraphics[width=\sizeImage\textwidth]{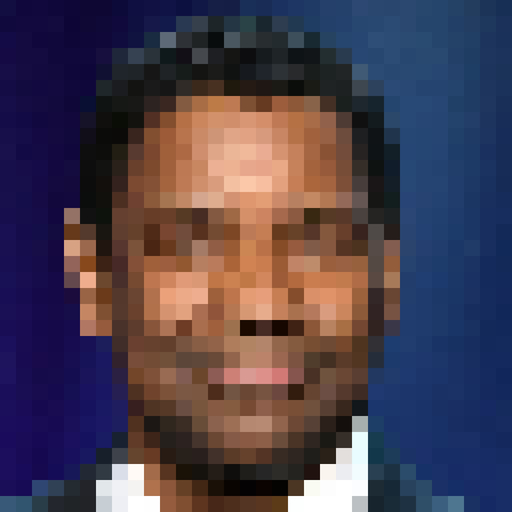}
    & \hspace{\sizeHorizontalSpace}
    \includegraphics[width=\sizeImage\textwidth]{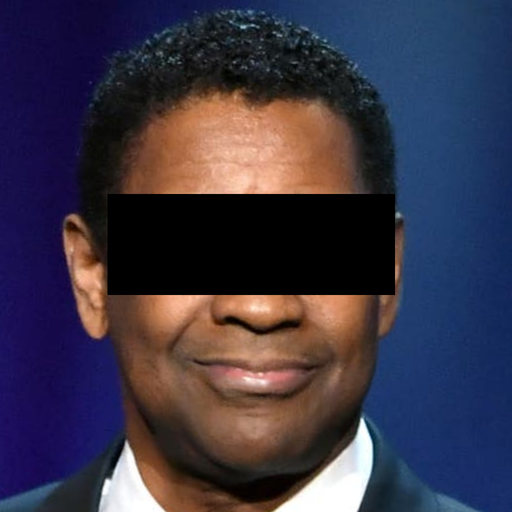}
    & \hspace{\sizeHorizontalSpace}
    \includegraphics[width=\sizeImage\textwidth]{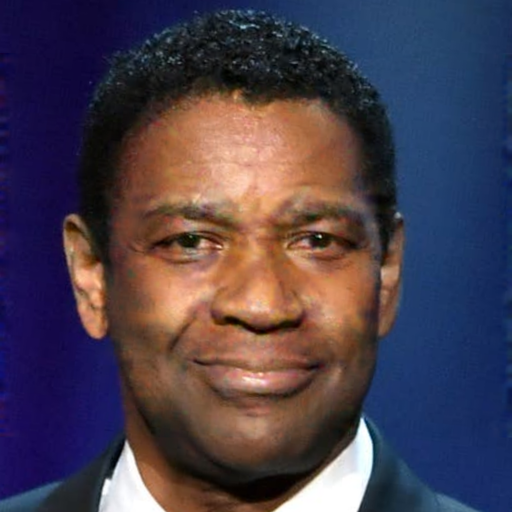}
    & \hspace{\sizeHorizontalSpace}
    \includegraphics[width=\sizeImage\textwidth]{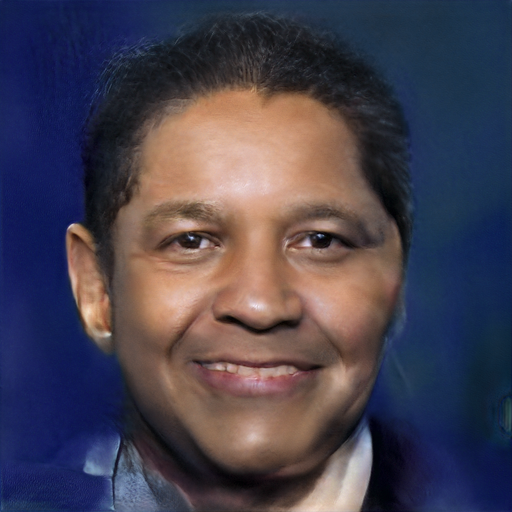}
    & \hspace{\sizeHorizontalSpace}
    \includegraphics[width=\sizeImage\textwidth]{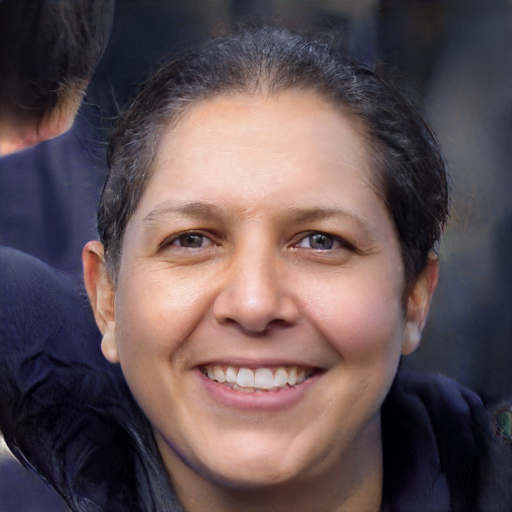}
\end{tabular}
\captionof{figure}{
    We propose a simple yet effective method for image obfuscation by blurring in latent space (i.e., DeepBlur).
    Comparing to existing methods (e.g., Gaussian blur, pixelation, masking, and adversarial noise),
    our approach preserves high perceptual quality
    while preventing unauthorized face recognition from both automatic systems and human adversaries.
}
\label{fig:teaser}
\end{center}

}]

\begin{abstract}\label{sec:abstract}
    There is a growing privacy concern 
    due to the popularity of social media and surveillance systems,
    along with advances in face recognition software.
    However, established image obfuscation techniques are either
    vulnerable to re-identification attacks by human or deep learning models,
    insufficient in preserving image fidelity,
    or too computationally intensive to be practical.
    To tackle these issues, we present DeepBlur, a simple yet effective method for image obfuscation
    by blurring in the latent space of an unconditionally pre-trained generative model that is able to synthesize photo-realistic facial images.
    We compare it with existing methods by efficiency and image quality,
    and evaluate against both state-of-the-art deep learning models
    and industrial products (e.g., Face++, Microsoft face service).
    Experiments show that our method
    produces high quality outputs
    and is the strongest defense for most test cases.
\end{abstract}

\begin{figure*}[t]
\centering
\includegraphics[width=0.9\textwidth]{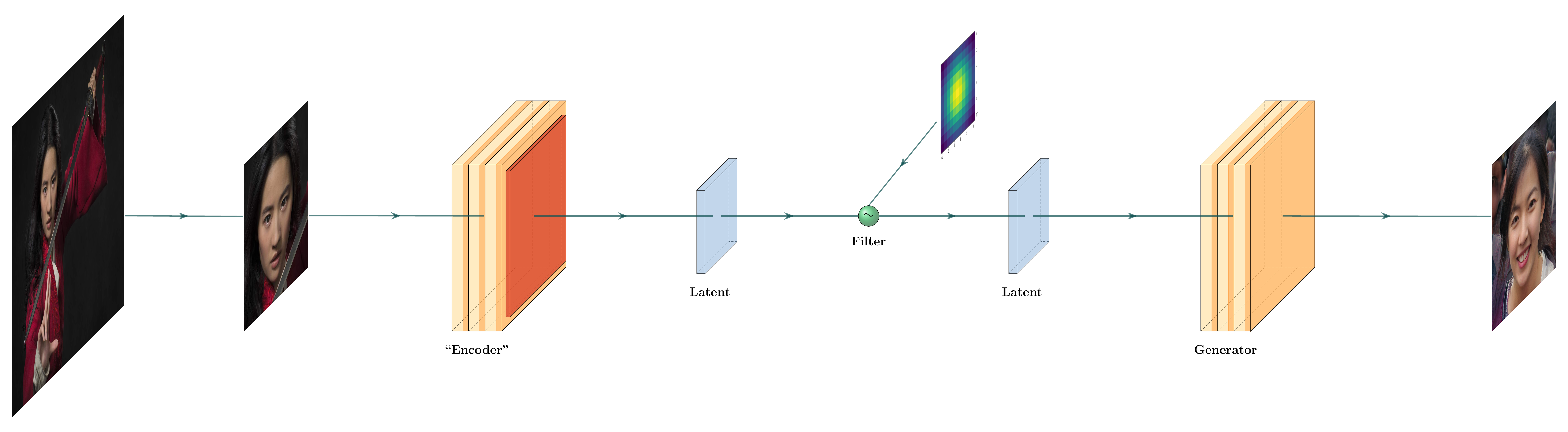}
\hspace{12pt}
\caption{
    DeepBlur overview.
    Given an arbitrary input, we first crop, rotate, and align the image,
    and then feed the aligned face to an optimization pipeline (see \cref{fig:latent_code_search}) 
    and obtain a latent representation that can synthesize almost identical faces with a generative neural network.
    Depending on the application scenario (e.g., the required level of obfuscation),
    we apply a low-pass filter with desired kernel size to the latent representation and provide the smoothed counterpart to the generative model again,
    which generates a ``deep blurred'' result.
}
\label{fig:method}
\end{figure*}

\section{Introduction}\label{sec:introduction}
    
Being in a digital era, we enjoy
the benefits of smartphones and cameras which facilitate learning and social connections.
In the meantime, however, billions of images being uploaded to public cloud servers every day,
introducing serious privacy concerns,
as an adversary may collect such data, identify ``persons of interest'' using either crowdsourcing or machine learning algorithms,
and secretly monitor our daily lives.
A recent New York Times article reveals that a private company collected over three billion online images
and trained a large model capable of recognizing millions of people without consent~\cite{hill2020secretive}.

With legal and privacy concerns, image obfuscation methods such as pixelation and blurring are often used to
protect sensitive information, e.g.,  human faces and confidential texts.
However, recent advances in  deep learning make these approaches less effective,
as it has been shown that blurred or pixelated facial images can be re-identified by
deep neural networks at high accuracy~\cite{mcpherson2016defeating}.
Moreover, image distorted by these methods usually are less visually pleasing (see examples in \cref{fig:teaser}).

More recently, new approaches such as adversarial perturbation \cite{tsipras2018robustness,wu2020making,shan2020fawkes}
and GAN-based image editing \cite{li2019anonymousnet,sun2018hybrid,hao2019utility} have been proposed to tackle these issues.
However, the former is subjective to the attack's deep learning model and generally fails when the model is unknown;
they also cannot (and are not supposed to) protect against human adversaries (e.g., crowdsourcing).
The latter, although may have better image quality and is capable of protecting against both human perception and automatic systems,
tends to be computationally expensive and may produce visible artifacts on synthesized faces,
especially when the GANs are conditionally trained~\cite{shen2020interfacegan}.

This leads to our motivation:
we need an image obfuscate method that is effective to protect against human and machine adversaries while preserving image quality,
yet simple enough, both conceptually and computationally, to be deployed in practice at a large scale.
Then we propose DeepBlur, a simple yet effective method for natural image obfuscation.
\Cref{fig:method} outlines the approach.

We argue that DeepBlur has following advantages over existing methods:
\begin{itemize}
    \item Compared to traditional methods (e.g., Gaussian blurring, pixelation, masking),
        DeepBlur is a stronger defense against deep learning-based recognition systems and is able to generate more visually pleasing results;
    \item Compared to adversarial perturbation-based methods,
        DeepBlur makes no assumption on the specific neural network or recognition system used by the attacker,
        and can defense against unauthorized recognition from both human crowdsourcing and automatic systems;
    \item Compared to GAN-based image editing methods (e.g., attribute editing, face inpainting and replacement), DeepBlur generally produces less artifacts (due to smoothing effect in latent space)
        and is more computationally friendly.
\end{itemize}
We will demonstrate these both qualitatively and quantitatively in the following sections.

The rest of the paper is organized as follows:
in \cref{sec:literature} we review recent advances in image privacy research and face manipulation techniques;
\cref{sec:method} formalizes the attack model and
explains the DeepBlur method, including our approach for latent representation search,  deep blurring, and image generation;
\cref{sec:experiment} details experiment settings and evaluation metrics, 
and show both qualitative and quantitative results;
\cref{sec:analysis} further discuss computational concerns and show interesting deep blur visual effects.
We concludes the paper in \cref{sec:conclusion}.

\section{Related Work}\label{sec:literature}

\paragraph{Privacy-Enhancing Techniques for Images.}

Classical methods such as pixelation and blurring (as shown in \cref{fig:teaser})
have been widely used to obfuscate facial images;
but, as mentioned earlier, they fail to defeat against modern facial recognition systems powered by deep learning
and often produce images that are not visually pleasing.
Methods include distorting images to make them unrecognizable~\cite{sun2018hybrid,wu2018privacy}, and
producing adversarial patches in the form of bright patterns printed on sweatshirts or signs, which prevent facial recognition algorithms from even registering their wearer as a person~\cite{thys2019fooling,wu2019making}.

However, these are targeted against facial recognition systems designed without regard to privacy protection, and could be subject to targeted re-identification attacks such as \cite{gross2005integrating,gross2006model,gross2009face}.
In 2005, Newton \etal \cite{newton2005preserving} introduced $k$-Same,
the first privacy-preserving algorithm in the context of image databases,
and Hao \etal \cite{hao2020robustness} demonstrated that it is more effective than canonical methods.
There is a trade-off between privacy and usability~\cite{li2019anonymousnet}
and Gross \etal \cite{gross2005integrating} introduced $k$-Same-Select to balance disclosure risk and classification accuracy.
Zhang \etal \cite{zhang2018privacy} further designed an ``obfuscate function'' that adds random noises to samples to hide sensitive information in the dataset while preserving model accuracy.
In 2018, Fan \cite{fan2018image} proposed an obfuscation method that satisfies
$\varepsilon$-differential privacy at pixel level,
yet its image quality is low and only protects privacy of the pixels instead of the person.
More recently, Li and Clifton \cite{li2021differentially} proposed
to manipulate image latent space
in a way that satisfies $\varepsilon$-differential privacy for the person
and produces photo-realistic images.

\begin{figure}[t]
\centering
\includegraphics[width=0.46\textwidth]{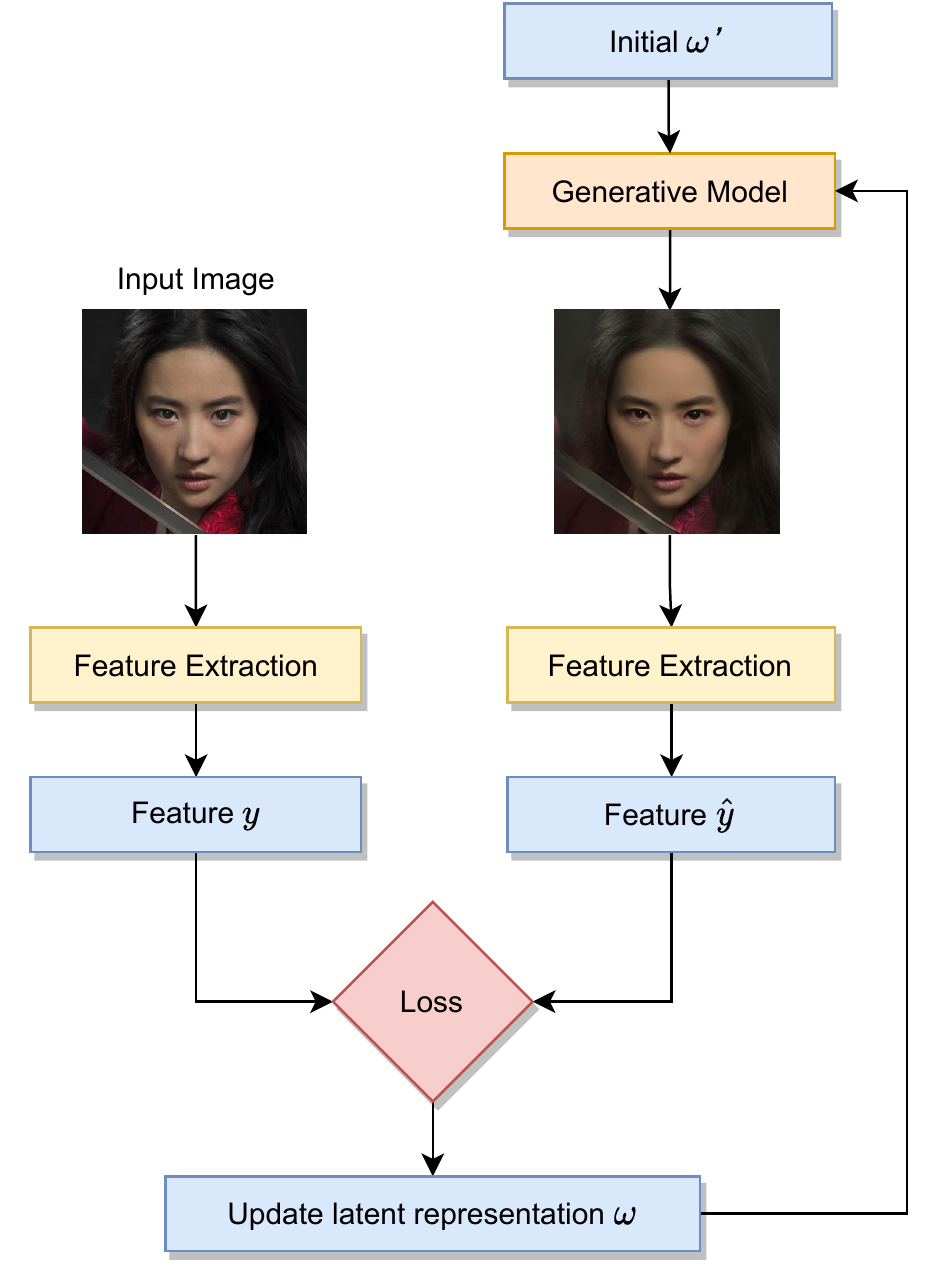}
\hspace{12pt}
\caption{
    A general framework of latent representation search.
}
\label{fig:latent_code_search}
\end{figure}

\begin{figure*}[t]
\begin{center}
\setlength{\tabcolsep}{.15em}
\begin{tabular}{ccccccc}
    $Input$ & $Step = 0$ & $Step = 1$ & $Step = 2$ & $Step = 3$ & $Step = 5$ & $Step = 10$ \\
    \includegraphics[width=\sizeImage\textwidth]{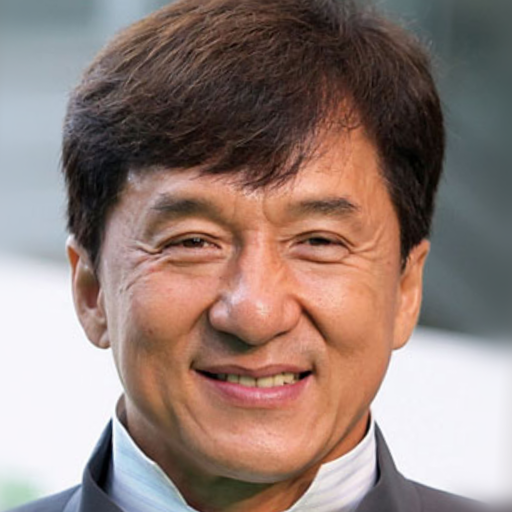}
    & \hspace{2pt}
    \includegraphics[width=\sizeImage\textwidth]{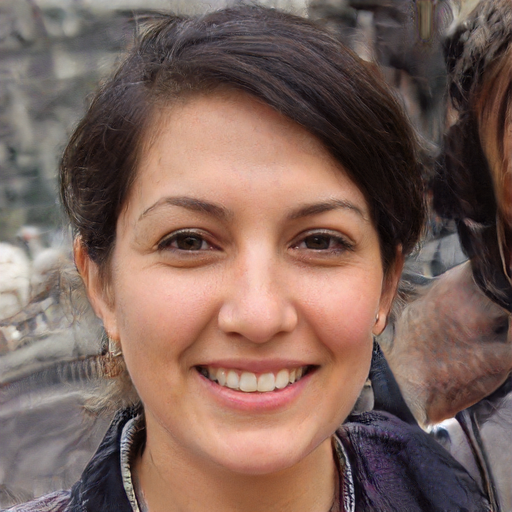}
    & \hspace{\sizeHorizontalSpace}
    \includegraphics[width=\sizeImage\textwidth]{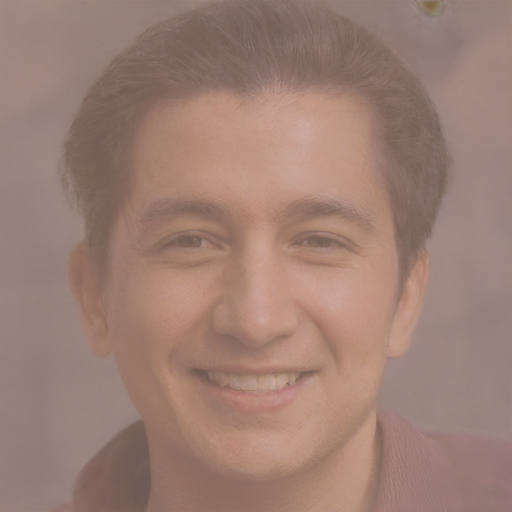}
    & \hspace{\sizeHorizontalSpace}
    \includegraphics[width=\sizeImage\textwidth]{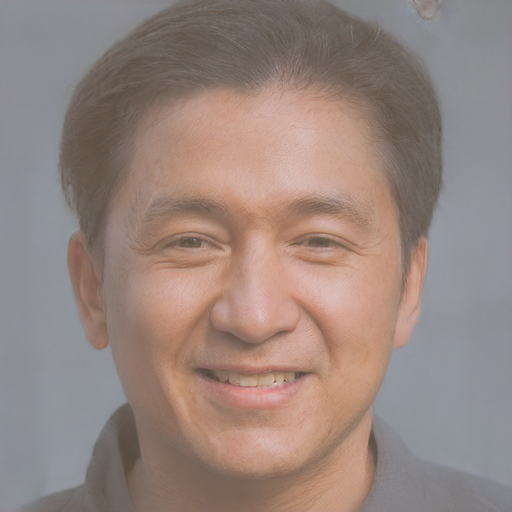}
    & \hspace{\sizeHorizontalSpace}
    \includegraphics[width=\sizeImage\textwidth]{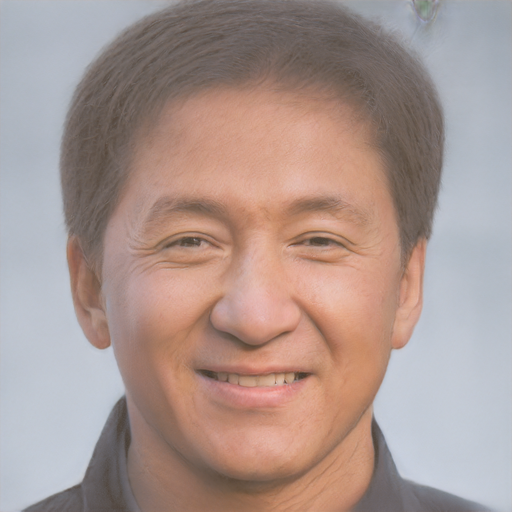}
    & \hspace{\sizeHorizontalSpace}
    \includegraphics[width=\sizeImage\textwidth]{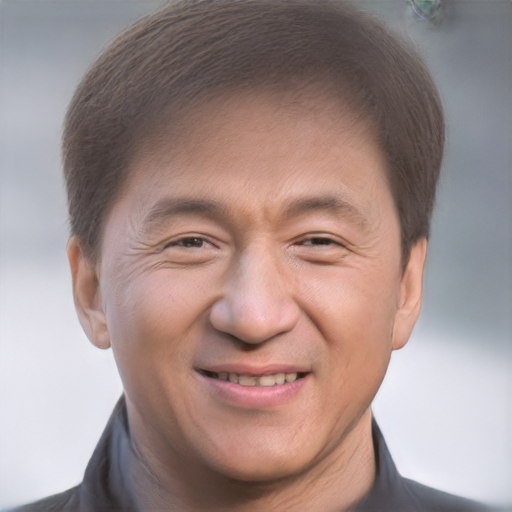}
    & \hspace{\sizeHorizontalSpace}
    \includegraphics[width=\sizeImage\textwidth]{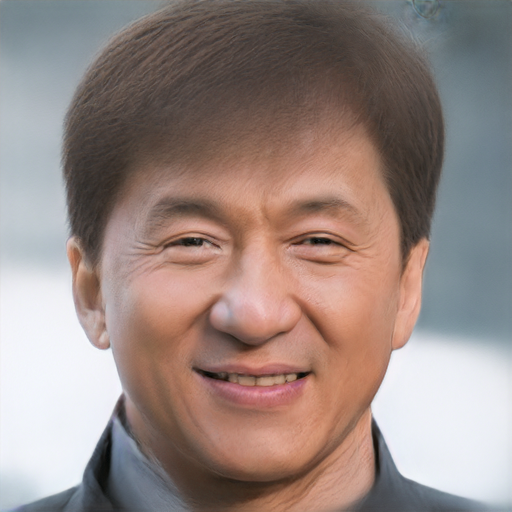}
    \\
    \includegraphics[width=\sizeImage\textwidth]{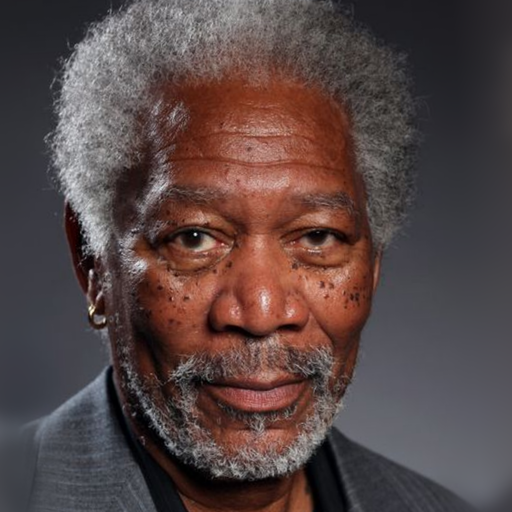}
    & \hspace{2pt}
    \includegraphics[width=\sizeImage\textwidth]{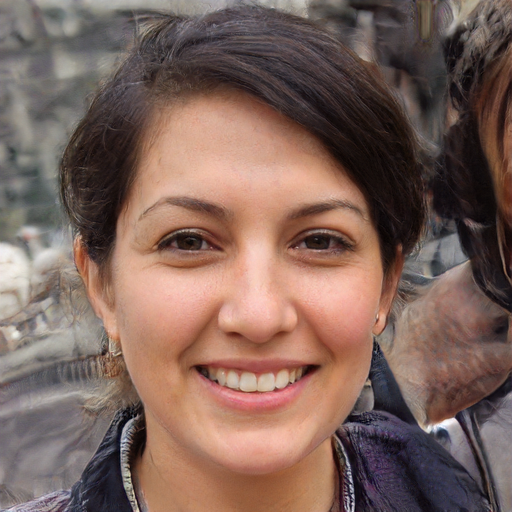}
    & \hspace{\sizeHorizontalSpace}
    \includegraphics[width=\sizeImage\textwidth]{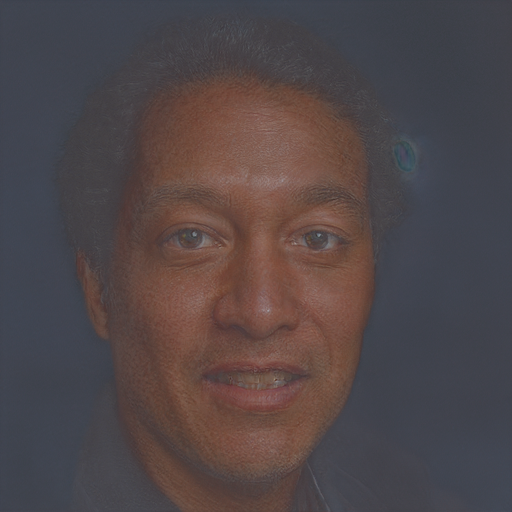}
    & \hspace{\sizeHorizontalSpace}
    \includegraphics[width=\sizeImage\textwidth]{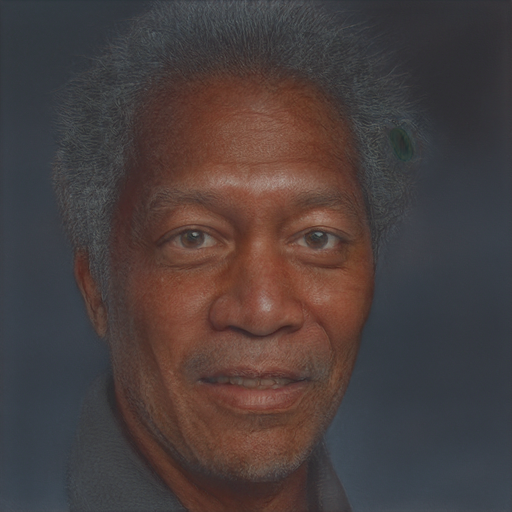}
    & \hspace{\sizeHorizontalSpace}
    \includegraphics[width=\sizeImage\textwidth]{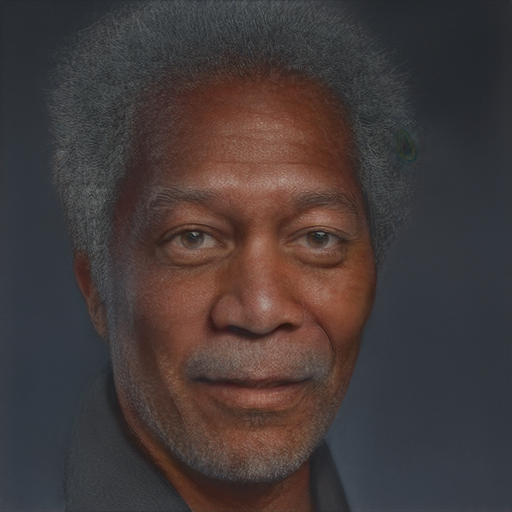}
    & \hspace{\sizeHorizontalSpace}
    \includegraphics[width=\sizeImage\textwidth]{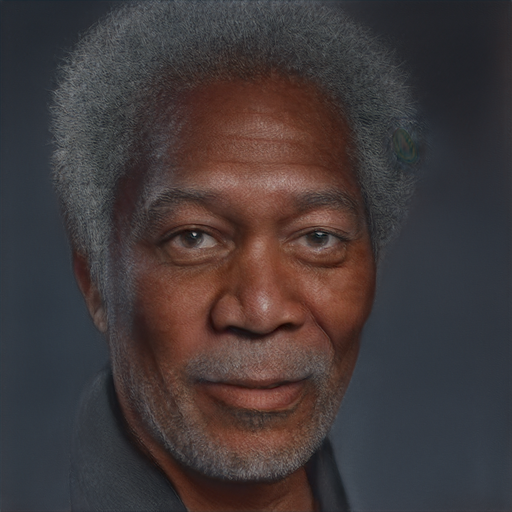}
    & \hspace{\sizeHorizontalSpace}
    \includegraphics[width=\sizeImage\textwidth]{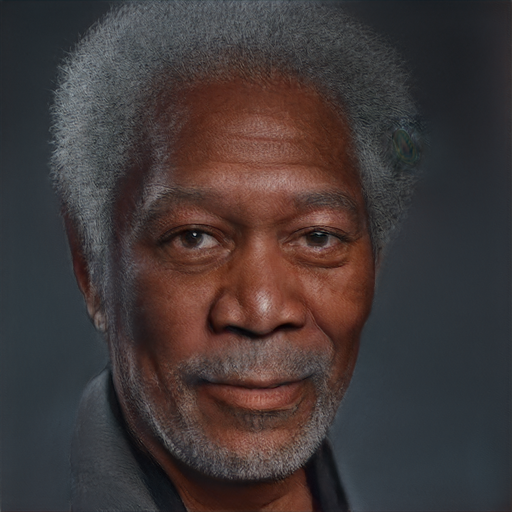}
\end{tabular}
\end{center}
\caption{Examples of latent representation search.
    As illustrated in \cref{fig:latent_code_search}, we formalize the task of finding the latent representation of images as an optimization problem.
    Starting from an ``average face'' of GAN (see \cref{fig:average_faces}),
    we use L-BFGS algorithm~\cite{fletcher2013practical} to find the latent representation that is able to minimize the loss between input image and the generated counterpart.
    The algorithm converges very quickly
    and usually a latent representation obtained after 10 steps
    can synthesize an image that looks reasonably close to the original.
}
\label{fig:optim_loss}
\end{figure*}

\paragraph{Facial Image Editing.}
Face analysis is an important topic in computer vision
with a wide range of real-world applications,
such as expression and attribute recognition \cite{shan2009facial,liu2018attributes}, face super-resolution \cite{kalarot2020component,jiang2021deep},
and virtual cosmetic enhancement \cite{liu2019understanding,liu2019face,li2019beauty}.
The task of facial image editing aims at manipulating facial attributes of a given image at the semantic level.
Current approaches include carefully designing loss functions~\cite{odena2017conditional,tran2017disentangled},
introducing additional attribute labels or features~\cite{lample2017fader,bao2018towards,shen2018facefeat},
and using special architectures to train new models~\cite{donahue2017semantically,shen2018faceid} .
However, the synthesized results by these conditionally trained models 
is incomparable to native unconditionally trained GANs,
such as PGGAN~\cite{karras2017progressive} and StyleGAN~\cite{karras2019style}.
Unlike AnonymousNet~\cite{li2019anonymousnet} and UP-GAN~\cite{hao2019utility} which use conditional GANs for image obfuscation,
DeepBlur leverages an unconditionally trained generative adversarial network
and varies its latent space to control image synthesis, which produces image outputs of higher quality
(see \cref{fig:teaser,fig:method,fig:latent_code_search,fig:optim_loss,fig:deepblur_demo}).

\paragraph{Latent Space Properties of GANs.}
Despite the great success of GANs in image synthesis and editing,
a full understanding of how semantics are encoded in their latent spaces is still missing.
A major issue is how we can project a measure (e.g., Euclidean distance) that we observe in semantic space
to its counterpart in latent space.
Literature usually treat the latent space as Riemannian manifold \cite{chen2017metrics,arvanitidis2017latent,kuhnel2018latent}.
Radford \etal \cite{radford2015unsupervised} and Upchurch \etal \cite{upchurch2017deep}
observed vector arithmetic properties in latent space
and analyzed the disentanglement of multiple semantics.
Studies in this domain are mostly empirical:
Jahanian \etal \cite{jahanian2019steerability} ``steered'' the latent space for camera motion and image scaling;
Yang \etal \cite{yang2019semantic} observed semantic hierarchy in scene synthesis models;
Bau \etal \cite{bau2018gan} found correspondences between intermediate layers of GANs and visual objects such as buildings and trees;
Shen \etal \cite{shen2020interfacegan} interpreted facial semantics by varying latent codes in the latent space.

\section{Preserving Image Privacy with DeepBlur}\label{sec:method}

In this section, we detail the DeepBlur method for image privacy preservation.
We first describe our assumptions for both users and attackers,
and define three threat models,
which explain
what we mean by
preserving image privacy.

\subsection{Assumptions and Threat Models}\label{sec:threat_model}

In our scenario, a user wants to upload images to a remote server,
where connections to the server or the server itself is compromised.
The goal of the user is to share high quality images with obfuscated identities,
with the hope that the user's identity will not be revealed even if an adversary has access to the images.

On the other side,
we assume that the attacker's goal is to build a powerful face recognition system
that can accurately identify a group of people.
We also assume that the attacker has unlimited computing resources
and can use external datasets to facilitate model training
(the external datasets exclude images from the users unless otherwise specified).
We say an obfuscation method is stronger than the other
if it has a lower identification accuracy by the attacker's recognition system.
Accordingly, we define three threat models.

\paragraph{Threat Model $T_1$.}
This scenario simulates the case that the attacker has prior information of the users
and tries to identify them from protected (i.e., obfuscated) images.
For example, a paparazzo obtains some sensitive personal photos
but the photos are obfuscated and he would like to know what celebrities are in the photos.
In other words, the paparazzo can train his model using all publicly available photos of celebrities
but the obfuscated photos were unseen.

\paragraph{Threat Model $T_2$.}
In this scenario, the attacker acquires a set of obfuscated images with identity labels
and tries to identify users from a group of original images.
For example, Eve has two classmates, Alice and Bob,
who have accounts in an anonymous dating website with selfies protected by image obfuscation techniques.
Eve downloads an obfuscated selfie and would like to train a model to predict
whose the selfie is.

\paragraph{Threat Model $T_3$.}
For this model, the attacker acquires obfuscated images with labels
and would like to associate the labels with another group of obfuscated images.
For example, an attacker successfully breaks into the server of \verb|victim.com|
which stores credentials and obfuscated photos of thousands of users.
Then he targets another website, 
\verb|vulnerable.com|,  whose preview mode shows obfsucated images of all users. 
To perform a credential stuffing attack
(i.e., use credentials on \verb|victim.com| to take over accounts on \verb|vulnerable.com|),
the attacker needs to train a model on the obfuscated images of \verb|victim.com|
and use it to label accounts on \verb|vulnerable.com|.

We will evaluate our method under these three settings
in \cref{sec:experiment}
and compare it with other obfuscation methods.

\subsection{The DeepBlur Method}\label{sec:deepblur}

From a high level perspective, DeepBlur applies a low-pass filter to the latent space of an input image
and then uses the blurred latent representation to synthesize the output.
\Cref{fig:method} shows an overview of DeepBlur,
including three essential steps: latent representation search (i.e., ``encoder''), deep blurring, and image generation,

\paragraph{Latent representation search.}
Given an arbitrary input image,
we first perform cropping and alignment,
and then feed it into a feature extractor (e.g., VGG16~\cite{simonyan2014very}) to obtain feature vector $y$.
In the meantime, we put its counterpart, an image synthesized by a generative model (e.g., StyleGAN~\cite{karras2019style})
with the same procedure and obtain feature vector $\hat{y}$.
After computing the loss between $y$ and $\hat{y}$,
we accordingly update the latent representation $\omega$, feed it back to the generator,
and repeat this process until the synthesized image is close enough to the original.
\Cref{fig:latent_code_search} illustrates the approach.

The reason that we don't use an autoencoder directly is that the latent representation obtained by such method
is hard to produce images close to the original and with quality comparable to ours (see \cref{fig:optim_loss});
also, with an efficient optimization algorithm, the proposed search method can converge quite quickly.
We will discuss this more in \cref{sec:analysis}.

\paragraph{Deep blurring.}

The two-dimensional Gaussian filter~\cite{haddad1991class} is defined as follows:
\begin{align}
    g(x, y) = \frac{1}{2\pi\sigma^2} e^{-\frac{x^2 + y^2}{2\sigma^2}},
\end{align}
where $x$ is the distance from the origin in $x$-axis,
$y$ is the distance from the origin in $y$-axis,
and $\sigma$ is the standard deviation of the Gaussian distribution.
From the previous step, we obtained $\omega$, the latent representation of the image.
Then we apply a filter on it,
and get the blurred representation,
\begin{equation}
    \omega' = g(\omega).
\end{equation}
Usually, $\omega$ is two-dimensional but the dimension may
vary depending on the specific generative model and layer in use.

\begin{figure*}[t]
\begin{center}
\setlength{\tabcolsep}{.15em}
\begin{tabular}{ccccccc}
    $Original$ & $\sigma = 0.25$ & $\sigma = 0.50$ & $\sigma = 0.75$ & $\sigma = 1.00$ & $\sigma = 1.50$ & $\sigma = 2.00$ \\
    \includegraphics[width=\sizeImage\textwidth]{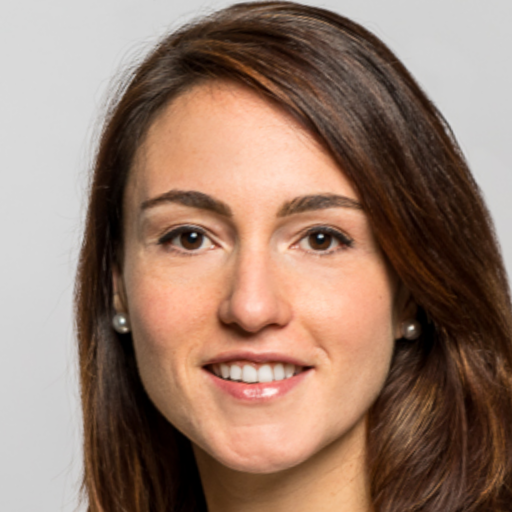}
    & ~
    \includegraphics[width=\sizeImage\textwidth]{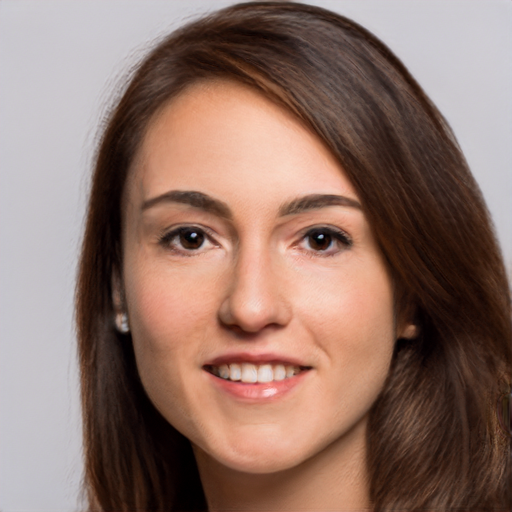}
    & \hspace{\sizeHorizontalSpace}
    \includegraphics[width=\sizeImage\textwidth]{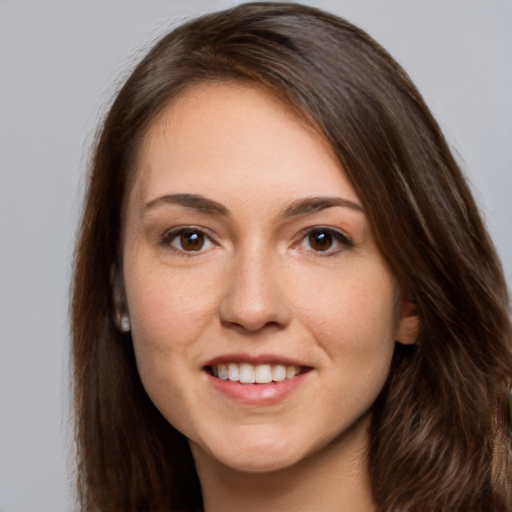}
    & \hspace{\sizeHorizontalSpace}
    \includegraphics[width=\sizeImage\textwidth]{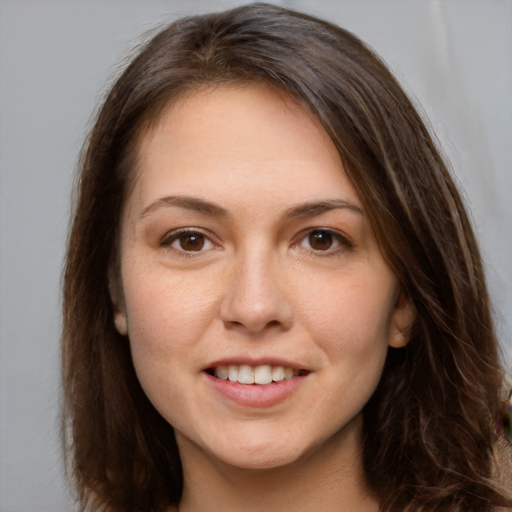}
    & \hspace{\sizeHorizontalSpace}
    \includegraphics[width=\sizeImage\textwidth]{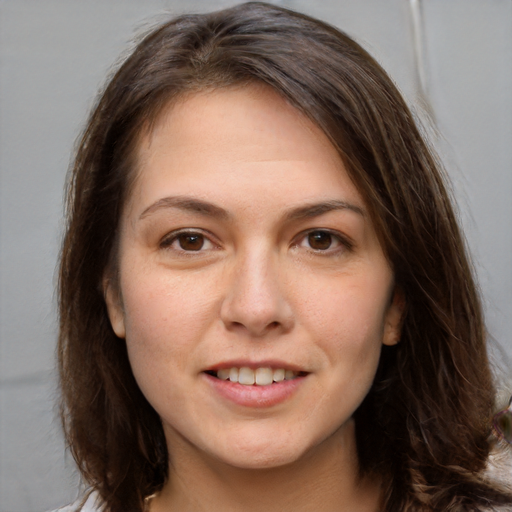}
    & \hspace{\sizeHorizontalSpace}
    \includegraphics[width=\sizeImage\textwidth]{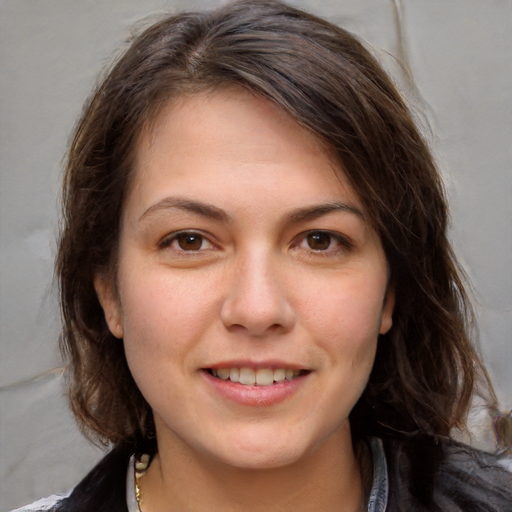}
    & \hspace{\sizeHorizontalSpace}
    \includegraphics[width=\sizeImage\textwidth]{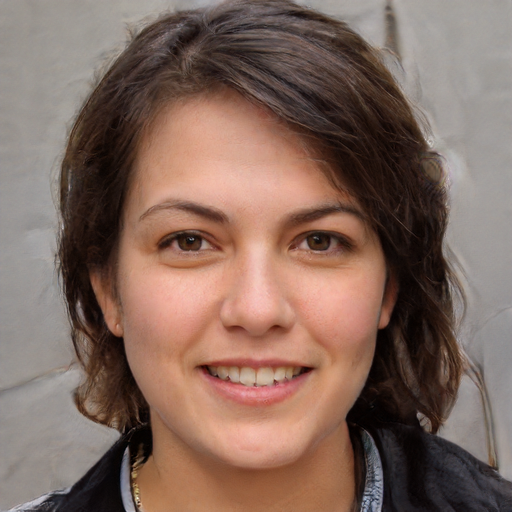}
    \\
    \includegraphics[width=\sizeImage\textwidth]{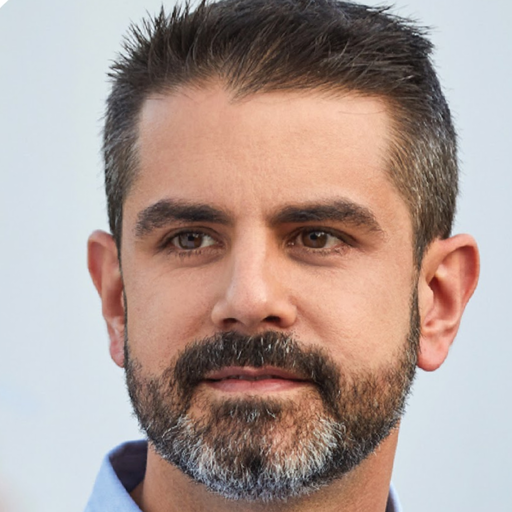}
    & ~
    \includegraphics[width=\sizeImage\textwidth]{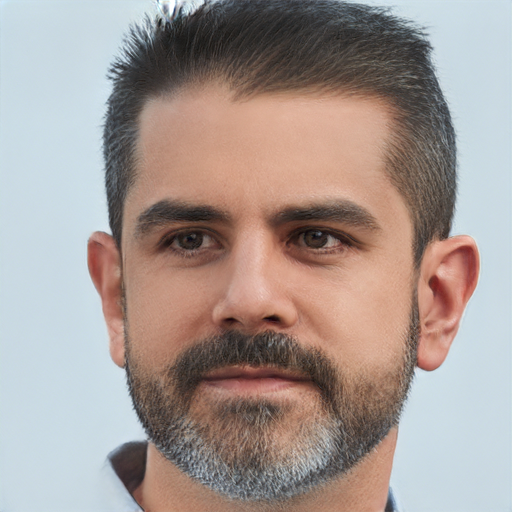}
    & \hspace{\sizeHorizontalSpace}
    \includegraphics[width=\sizeImage\textwidth]{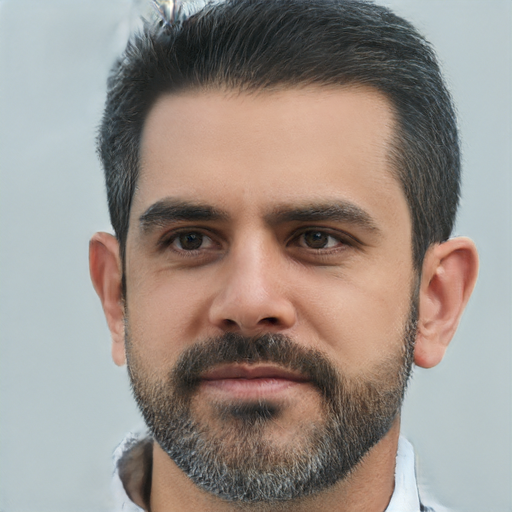}
    & \hspace{\sizeHorizontalSpace}
    \includegraphics[width=\sizeImage\textwidth]{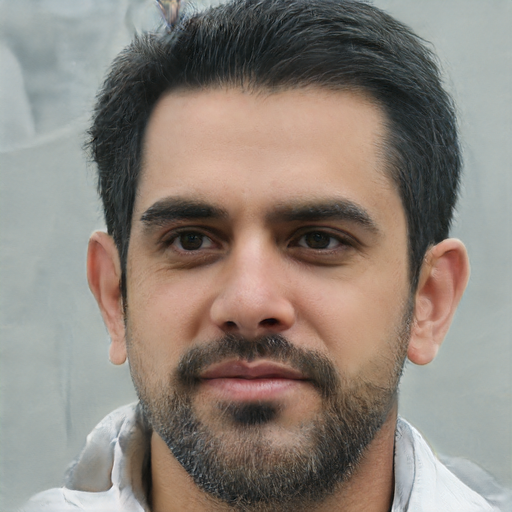}
    & \hspace{\sizeHorizontalSpace}
    \includegraphics[width=\sizeImage\textwidth]{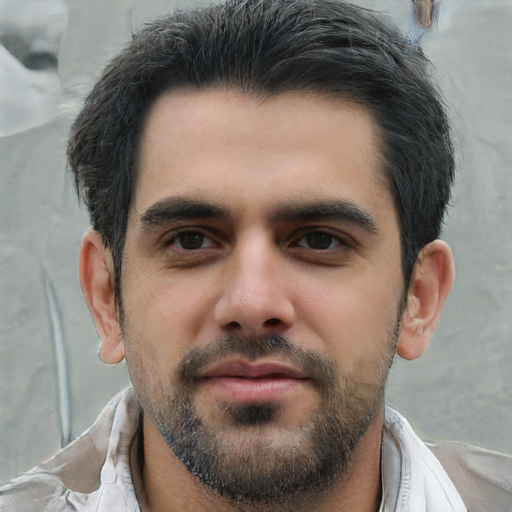}
    & \hspace{\sizeHorizontalSpace}
    \includegraphics[width=\sizeImage\textwidth]{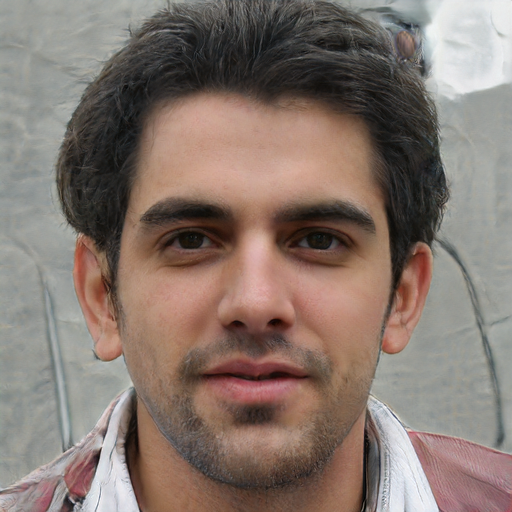}
    & \hspace{\sizeHorizontalSpace}
    \includegraphics[width=\sizeImage\textwidth]{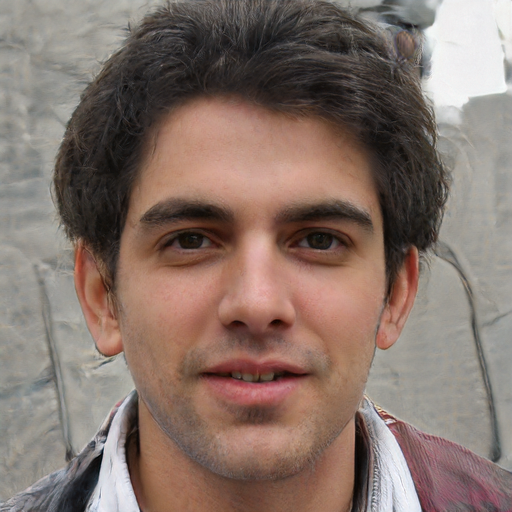}
    \\
    \includegraphics[width=\sizeImage\textwidth]{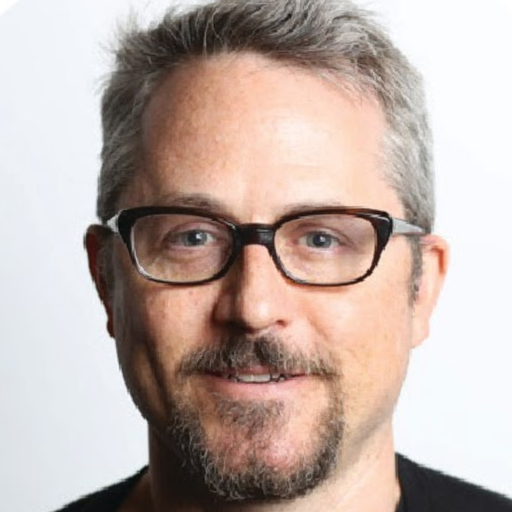}
    & ~
    \includegraphics[width=\sizeImage\textwidth]{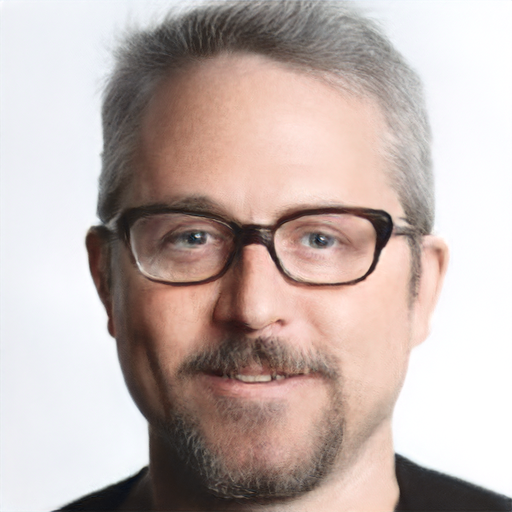}
    & \hspace{\sizeHorizontalSpace}
    \includegraphics[width=\sizeImage\textwidth]{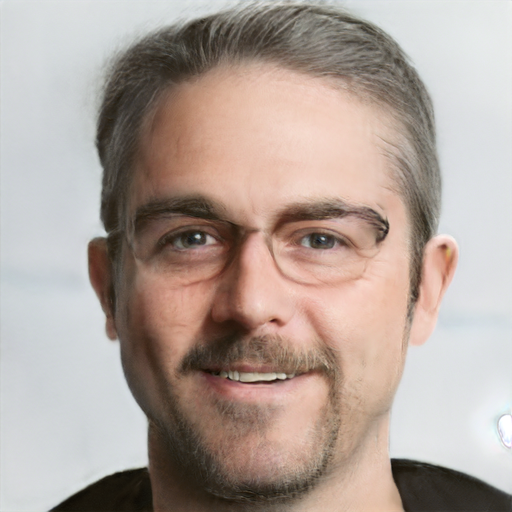}
    & \hspace{\sizeHorizontalSpace}
    \includegraphics[width=\sizeImage\textwidth]{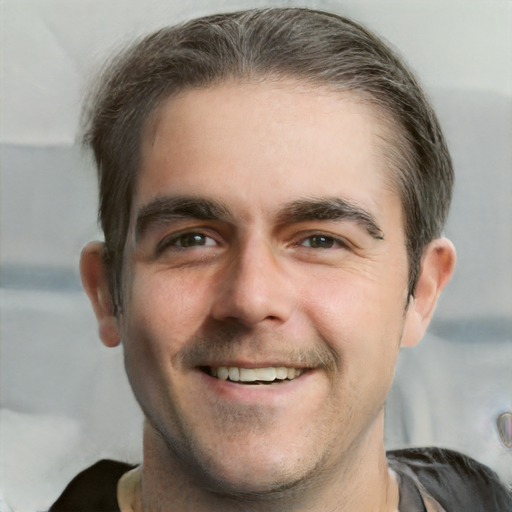}
    & \hspace{\sizeHorizontalSpace}
    \includegraphics[width=\sizeImage\textwidth]{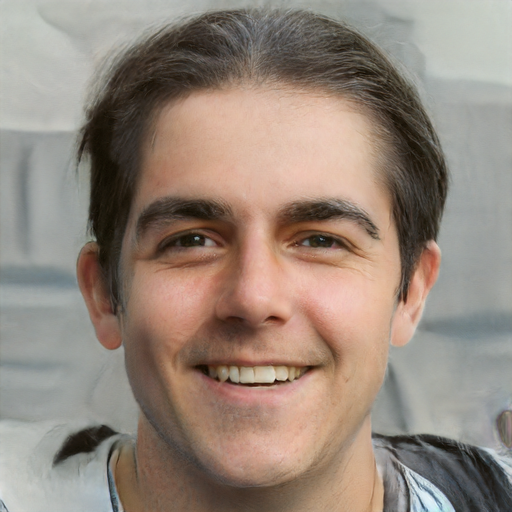}
    & \hspace{\sizeHorizontalSpace}
    \includegraphics[width=\sizeImage\textwidth]{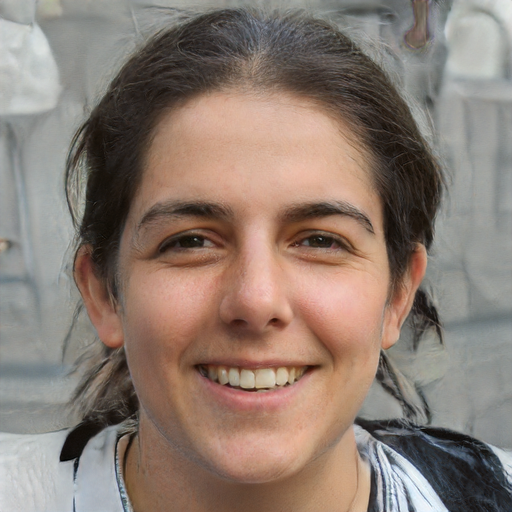}
    & \hspace{\sizeHorizontalSpace}
    \includegraphics[width=\sizeImage\textwidth]{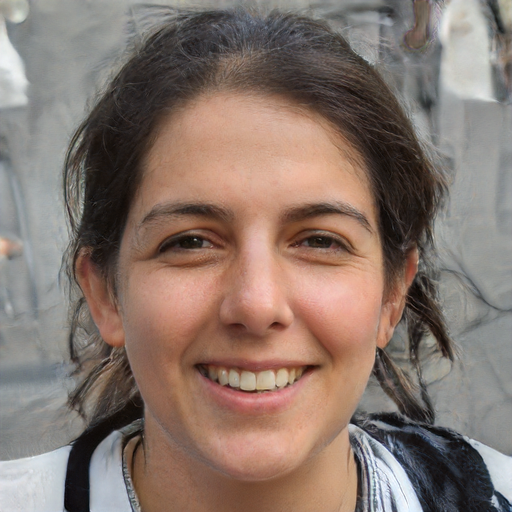}
    \\
    \includegraphics[width=\sizeImage\textwidth]{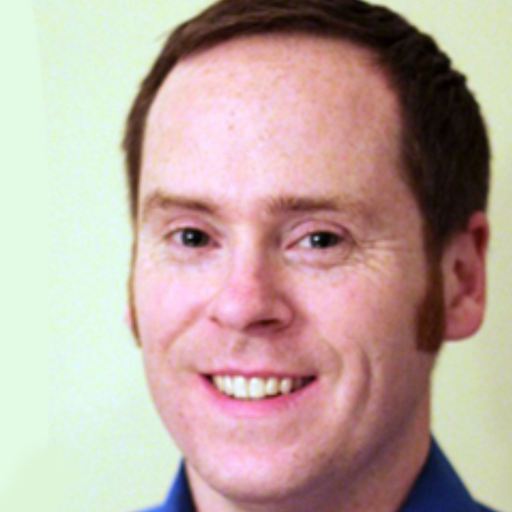}
    & ~
    \includegraphics[width=\sizeImage\textwidth]{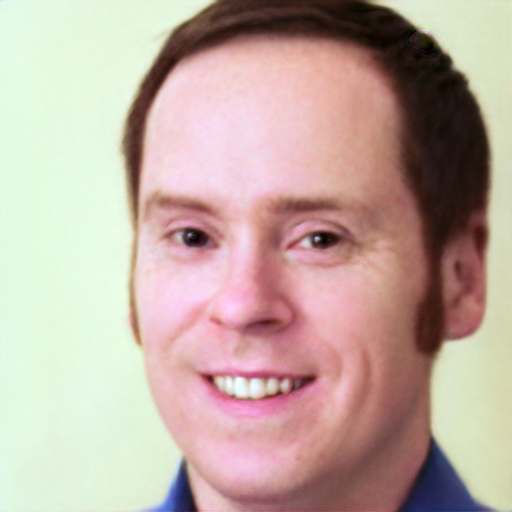}
    & \hspace{\sizeHorizontalSpace}
    \includegraphics[width=\sizeImage\textwidth]{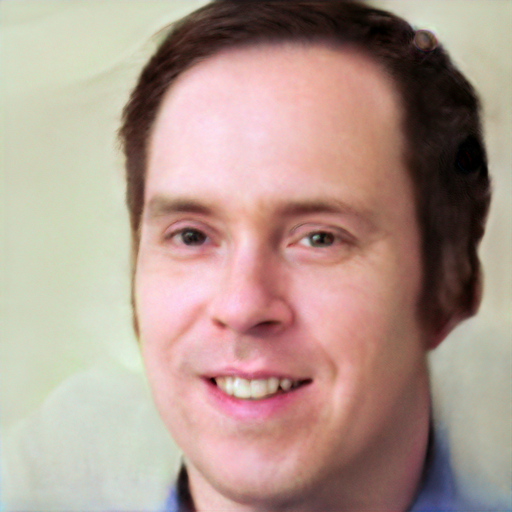}
    & \hspace{\sizeHorizontalSpace}
    \includegraphics[width=\sizeImage\textwidth]{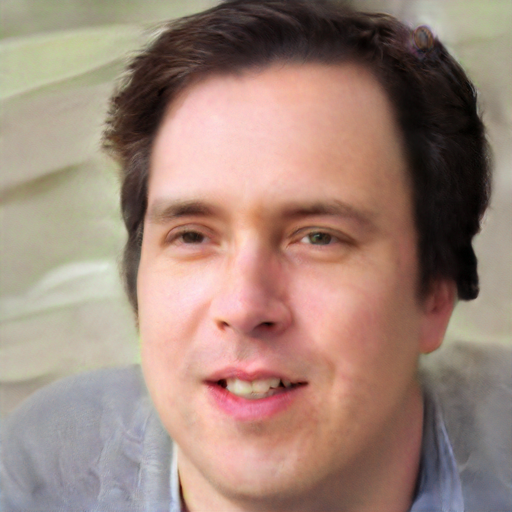}
    & \hspace{\sizeHorizontalSpace}
    \includegraphics[width=\sizeImage\textwidth]{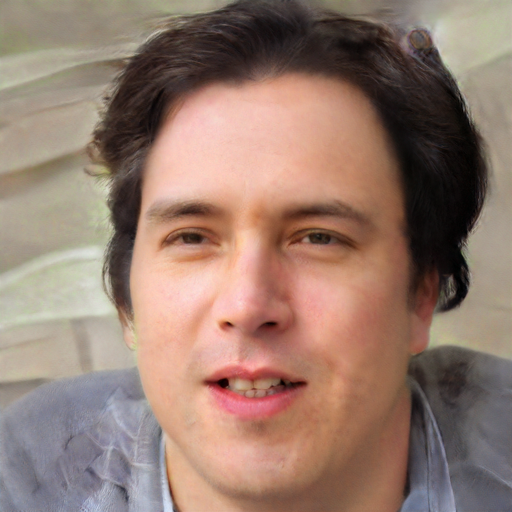}
    & \hspace{\sizeHorizontalSpace}
    \includegraphics[width=\sizeImage\textwidth]{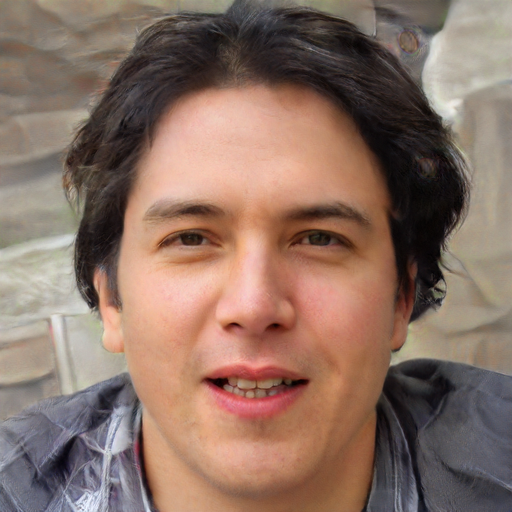}
    & \hspace{\sizeHorizontalSpace}
    \includegraphics[width=\sizeImage\textwidth]{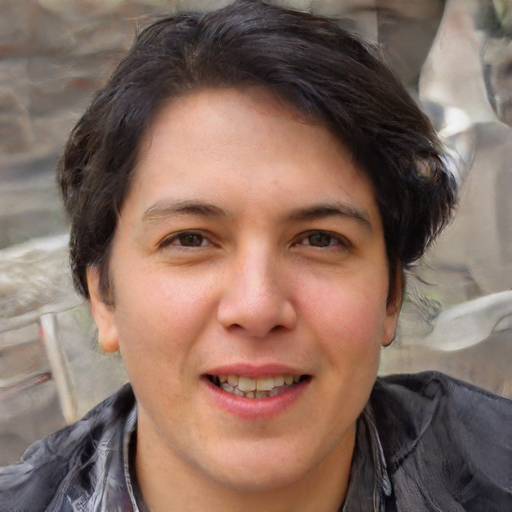}
\end{tabular}
\end{center}
\caption{
    Examples of deep blurred images.
    Given an arbitrary facial images, the DeepBlur method is able to obfuscate the identity while preserving high visual fidelity,
    and the identity distance monotonically increases as $\sigma$ getting larger.
    Note that the original images are from CVPR'21 Media Forensics Workshop committee
    and were not visible to the pre-trained generative model (see \cref{sec:generative_model})
    during training.
}
\label{fig:deepblur_demo}
\end{figure*}

\paragraph{Image generation from latent representation.}\label{sec:generative_model}
After the above deep blurring step, we obtained a latent representation of the image that was smoothed in its latent space.
The process of image generation using GAN is to feed the latent values into a specific layer in the generator.
In our case, we feed the blurred latent representation $\omega$ to the unconditionally pre-trained generative model in the searching step (see \cref{fig:latent_code_search}).
By changing the kernel size (i.e., $\sigma$), 
we can adjust the output image to a desired level of obfuscation.
\Cref{fig:deepblur_demo} demonstrates deep blurred images with various kernel sizes.

\section{Experiment}\label{sec:experiment}
Compared to existing methods, DeepBlur shows convincing performance in terms of
both effectiveness against adversarial facial recognition systems
and the quality of synthesized images.
In this section, we first introduce the datasets and experimental settings in \cref{sec:dataset},
and assess image quality in \cref{sec:metrics}.
We then evaluate obfuscation methods in \cref{sec:evaluation} under different attack settings.

\subsection{Datasets}\label{sec:dataset}
In our study, we mainly use two datasets:
FlickrFaces-HQ (FFHQ) \cite{karras2019style} and CelebFaces Attributes (CelebA) \cite{liu2015faceattributes}.

The former, FFHQ, consists of 70,000 high-resolution (i.e., 1024 $\times$ 1024) images,
covering a wide spectrum of faces with various ages, ethnicities, and image backgrounds.
It was collected by researchers in NVIDIA from Flickr.
The style-based generator that we use in the latent search step and image generation step (see \cref{fig:method})
was trained in this dataset.

The latter, CelebA, is a large-scale human face dataset which contains more than 200,000 images from over 10,000 celebrities (i.e., different identities).
We use the dataset to evaluate and compare our approach with others.
For proof of concept, instead of using the entire dataset, we select a subset of 100 identities and 10 images for each,
and split the 10 images as 7, 1, and 2, for training, validation, and testing, respectively.
Note that necessary preprocessing procedures (e.g., face alignment) are performed for all images before running the experiments.

\begin{figure*}[t]
\centering
\includegraphics[width=0.31\textwidth]{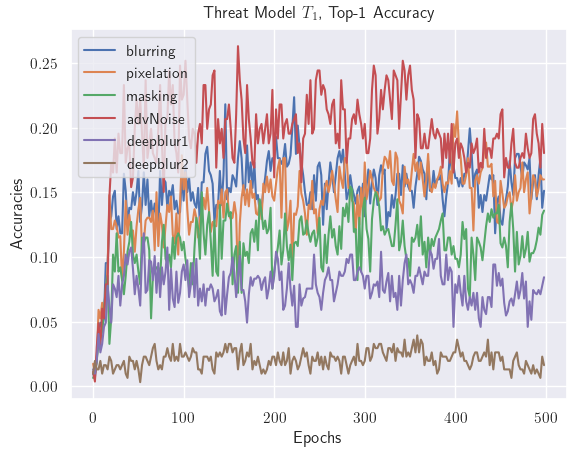}
\includegraphics[width=0.31\textwidth]{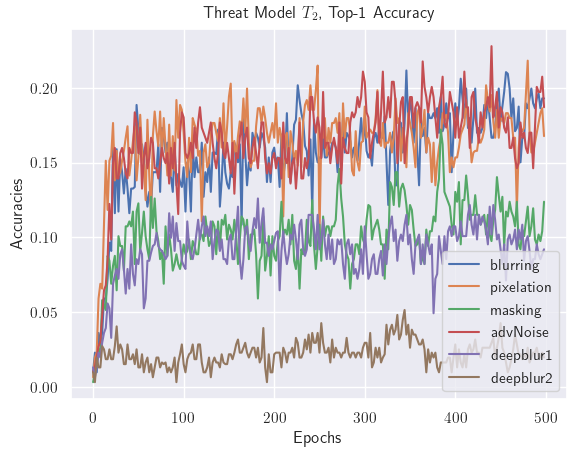}
\includegraphics[width=0.31\textwidth]{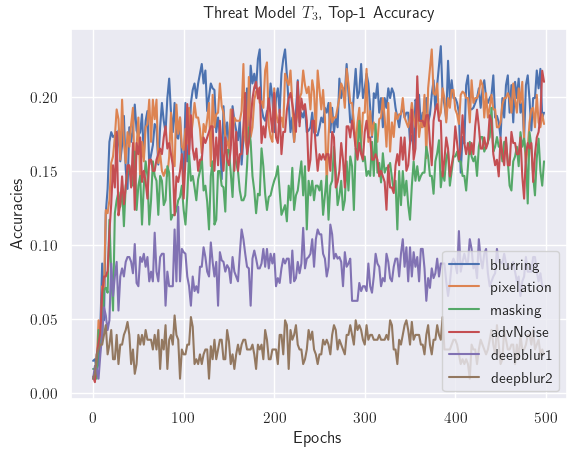}
\\
\includegraphics[width=0.31\textwidth]{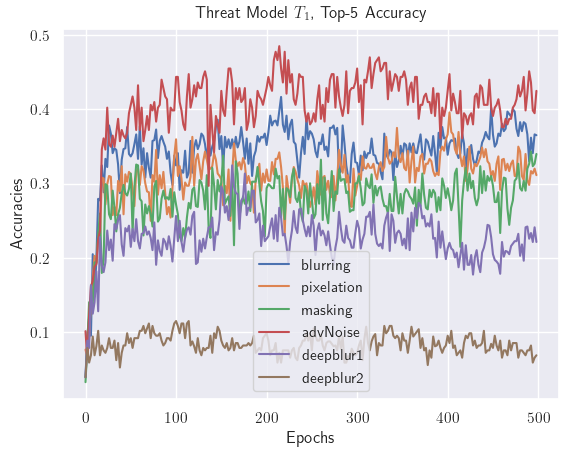}
\includegraphics[width=0.31\textwidth]{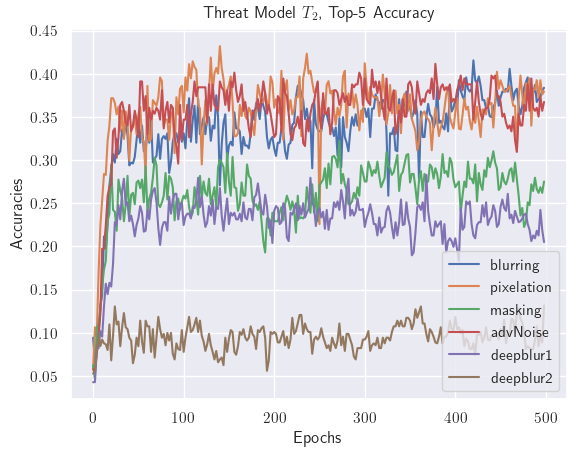}
\includegraphics[width=0.31\textwidth]{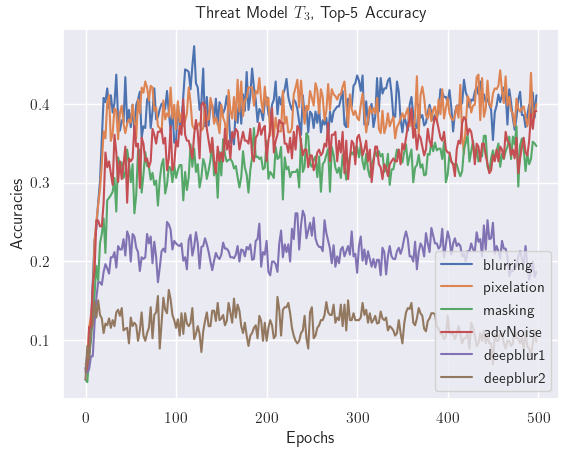}
\caption{
    Comparison of obfuscation methods.
    Above, we show the test result from a VGG16 network~\cite{simonyan2014very} trained after 500 epochs under different settings
    (top row and bottom row report top-1 accuracies and top-5 accuracies respectively,
    and each column corresponds to a threat model specified in \cref{sec:threat_model}).
    We compare Gaussian blurring on pixels, pixelation, masking, adversarial noise by Fawkes~\cite{shan2020fawkes},
    and our method with two settings ($\sigma=0.5$ and $\sigma=1.0$).
    Note that our settings are the same as shown in \cref{fig:teaser,fig:deepblur_demo}.
    More test results from other face recognition systems (e.g., commercial APIs) can be found in \cref{tab:accuracies_merged}.
}
\label{fig:accuracies}
\end{figure*}

\subsection{Image Quality Assessment}\label{sec:metrics}
Structural Similarity Index Measure (SSIM) and 
Fr\'echet Inception distance (FID) are commonly used to measure image quality
in terms of similarity and perceptual distance.
\begin{definition}
Given a reference image and a test image, the PSNR (in dB) between the two images is defined as
\begin{align}
\text{PSNR} = 10 \cdot \log_{10} ( \frac{\text{MAX}^2_\text{I}}{\text{MSE}} ),
\end{align}
where $\text{MAX}_\text{I}$ is the maximum possible pixel value of the image (typically 255)~\cite{hore2010image}. 
\end{definition} 
\begin{definition}
\begin{align}
    \text{FID} = ||\mu_r - \mu_g||^2 + \text{Tr} (\Sigma_r + \Sigma_g - 2 (\Sigma_r \Sigma_g)^{1/2}),
\end{align}
where $X_r \sim \mathcal{N} (\mu_r, \Sigma_r)$ and $X_g \sim \mathcal{N} (\mu_g, \Sigma_g)$ are activations of Inception-v3 pool3 layer for real and generated samples, respectively.
\end{definition}

\begin{table}[b]
\begin{center}
\resizebox{0.33\textwidth}{!}{%
\begin{tabular}{@{}lccc@{}}
\toprule
                & SSIM           & MS-SSIM        & FID              \\ \midrule
Blurring        & 0.858          & 0.814          & 60.770           \\
Pixelation      & 0.759          & 0.737          & \textbf{264.946} \\
Masking         & 0.873          & 0.889          & 72.417           \\
AdvNoise        & 0.482          & 0.393          & 42.764           \\
Ours$^\dagger$  & 0.467          & 0.380          & 207.473          \\
Ours$^\ddagger$ & \textbf{0.430} & \textbf{0.349} & 231.115          \\ \bottomrule
\end{tabular}
}
\end{center}
\caption{Measures of identity distance.
    We use SSIM and MS-SSIM~\cite{wang2003multiscale} to measure the similarity between original image and the obfuscated counterpart,
    and use Fr\'echet inception distance (FID)~\cite{heusel2017gans} for distance between the original and obfuscated identities.
    For our method, we set the $\sigma$ values as 2 and 5 respectively.
}
\label{tab:metrics}
\end{table}

A larger FID means the generated result is further away from the original
in the identity space (as measured by the Fr\'echet distance between the two distributions),
and a larger value of SSIM or MS-SSIM implies two images are structurally more similar.
\Cref{tab:metrics} compares obfuscation methods in terms of SSIM, MS-SSIM, and FID.
Results imply that DeepBlur well preserves structural similarity and quality of images while the generated identities are far from the original,
which aligns with our observation in \cref{fig:teaser} and \cref{fig:deepblur_demo}.
Note that SSIM may fail to capture nuances of human perception~\cite{zhang2018unreasonable}
and a smaller or larger value of the metrics  does not necessarily imply higher or lower image quality.
Thus, we only use the measurements for reference.

\begin{table*}[t]
\begin{center}
\resizebox{\textwidth}{!}{%
\begin{tabular}{@{}ccccccccccccc@{}}
\toprule
                    & \multicolumn{4}{c}{Threat Model $T_1$}                            & \multicolumn{4}{c}{Threat Model $T_2$}                            & \multicolumn{4}{c}{Threat Model $T_3$}                            \\ \cmidrule(l){2-13}
                    & VGG19          & ResNet18       & Face++         & Azure          & VGG19          & ResNet18       & Face++         & Azure          & VGG19          & ResNet18       & Face++         & Azure          \\ \midrule
Original            & 0.208          & 0.421          & 0.973          & 0.935          & 0.208          & 0.421          & 0.973          & 0.935          & 0.208          & 0.421          & 0.973          & 0.935          \\
Pixelation          & 0.160          & 0.305          & 0.458          & 0.255          & 0.168          & 0.373          & \textbf{0.116} & 0.391          & 0.182          & 0.155          & 0.292          & 0.653          \\
Blurring            & 0.151          & 0.157          & 0.922          & 0.592          & 0.193          & 0.056          & 0.942          & 0.684          & 0.190          & 0.397          & 0.912          & 0.871          \\
Masking             & 0.136          & 0.072          & 0.614          & 0.289          & 0.124          & 0.223          & 0.646          & 0.201          & 0.157          & 0.114          & 0.537          & \textbf{0.051} \\
AdvNoise            & 0.180          & 0.366          & 0.951          & 0.765          & 0.187          & 0.314          & 0.908          & 0.878          & 0.211          & 0.256          & 0.910          & 0.760          \\
DeepBlur$^\dagger$  & 0.084          & 0.209          & 0.683          & 0.330          & 0.092          & 0.213          & 0.908          & 0.500          & 0.072          & 0.225          & 0.541          & 0.460          \\
DeepBlur$^\ddagger$ & \textbf{0.016} & \textbf{0.043} & \textbf{0.060} & \textbf{0.000} & \textbf{0.020} & \textbf{0.020} & 0.180          & \textbf{0.000} & \textbf{0.026} & \textbf{0.032} & \textbf{0.070} & 0.126          \\ \bottomrule
\end{tabular}%
}
\end{center}
\caption{
    Comparison of obfuscation methods.
    We evaluate obfuscation methods by top-1 accuracies 
    of four face recognition systems (i.e., VGG16, ResNet18, Face++, and Microsoft Azure Face API) under different threat model settings.
    DeepBlur$^\dagger$ and DeepBlur$^\ddagger$ correspond to $\sigma = 0.5$ and $\sigma = 1.0$ respectively,
    and experiments show that the later one is the strongest at most times.
    Note that the experiments use the same settings as in \cref{fig:teaser,fig:deepblur_demo},
    and ``original'' means the original image without any obfuscation.
}
\label{tab:accuracies_merged}
\end{table*}

\begin{figure*}[t]
\centering
\includegraphics[width=0.35\textwidth]{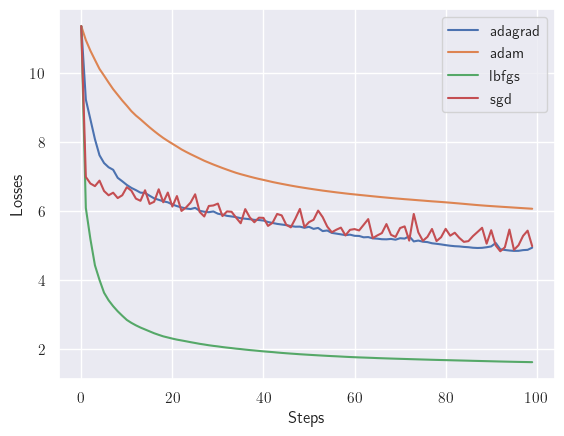}
\hspace{16pt}
\includegraphics[width=0.35\textwidth]{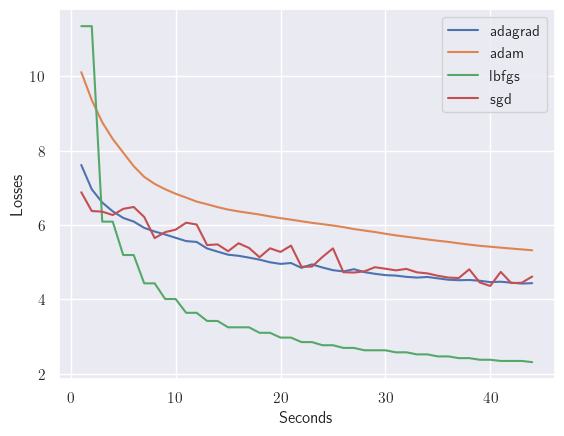}
\caption{
    We evaluate four canonical optimization methods for latent representation search,
    including Adam~\cite{kingma2014adam}, AdaGrad~\cite{duchi2011adaptive},
    SGD with momentum~\cite{qian1999momentum}, and L-BFGS~\cite{fletcher2013practical},
    and find that L-BFGS outperforms the rest in terms of efficiency.
    Note that a very accurate image latent representation isn't necessary for our task.
    Thus, we weight speed over precision and argue that L-BFGS is a good fit here.
}
\label{fig:optim_method_compare}
\end{figure*}

\subsection{Evaluation of Obfuscation Methods}\label{sec:evaluation}

We evaluate the obfuscation methods by attacking them under different threat model settings
(i.e., $T_1$, $T_2$, and $T_3$ as specified in \cref{sec:threat_model})
and with both canonical deep learning models
and commercial face recognition systems.

The task of face recognition is essentially a classification problem.
We first attack the obfuscation methods using VGG16 \cite{simonyan2014very}
which consists of 13 convolutional layers and 3 linear layers, and was the runner-up of ImageNet Large Scale Visual Recognition Competition (ILSVC) in 2014.
We also use ResNet18 \cite{he2016identity}, the winner of the ILSVC 2015 challenge, to simulate the attack scenarios, which has 17 convolutional layers with skip connections and 1 linear layer and uses the pre-activation residual unit.

With years of development in deep learning and face recognition techniques,
canonical models such VGG and ResNet may not reflect the real privacy threat today.
Thus, we also attack the obfuscation methods with commercial grade face recognition systems,
including  Microsoft Azure Face API \cite{azureface} and Face++ \cite{facepp}.
Microsoft Azure Face API is a part of Microsoft's cognitive services which provide various machine cognition algorithms
from face detection to cluster similar faces.
For their face identification API, a user can provide a set of mappings of identities and images for training
and query the identity of an image out of the training set.
The service returns the best matching along with confidence level among the given user pool.
Face++ provides similar API service for face identification where a user can enter annotated image data, train a model, and query the identity of the unseen.

\Cref{fig:accuracies} shows the experimental results under three threat model settings (i.e., $T_1$, $T_2$, and $T_3$ as specified in \cref{sec:threat_model})
and reports both top-1 and top-5 accuracies.
In the accuracy versus epoches plot,
the attacker (i.e., VGG16) achieves the lowest accuracy with DeepBlur$^\ddagger$ (i.e., $\sigma=1.0$),
meaning it is the strongest defense among all compared methods.
As mentioned in \cref{sec:dataset}, the test dataset has 100 identities.
Thus, the expected accuracy for a random guess is 0.01,
which is close to what this face recognition model can get with our approach.
Note that $\sigma = 1.0$ isn't an unreasonably large value as it still preserves high image quality
and certain facial semantics from the original (see \cref{fig:deepblur_demo} and ours$^\ddagger$ in \cref{fig:teaser}).
\Cref{tab:accuracies_merged} lists the results for all four face recognition methods,
showcasing that DeepBlur is the strongest method for most of the tests.

\section{Analysis}\label{sec:analysis}
In this section, we extend our analysis of DeepBlur,
and further discuss its computational efficiency in \cref{sec:complexity}
and show empirical results in \cref{sec:deepblur_effects} that may explain its superior visual quality
and tricks for fast convergence in latent representation search.

\subsection{Computational Efficiency of DeepBlur}\label{sec:complexity}

As shown in \cref{fig:method}, our approach has three main components:
latent representation search, deep blurring, and image generation.
It is trivial that the first is the most computationally expensive step,
as the second step is nothing more than
linear filtering
and the third step
only takes one forward pass in the pre-trained generative model

We formalize the searching step as an optimization problem in \cref{sec:deepblur}.
and use derivative information to update the latent representation.
Due to computational efficiency concerns,
we mainly investigate two types of optimization algorithms:
\begin{figure}[t]
\begin{center}
\begin{minipage}[b]{0.11\textwidth}\centering
    \includegraphics[width=\textwidth]{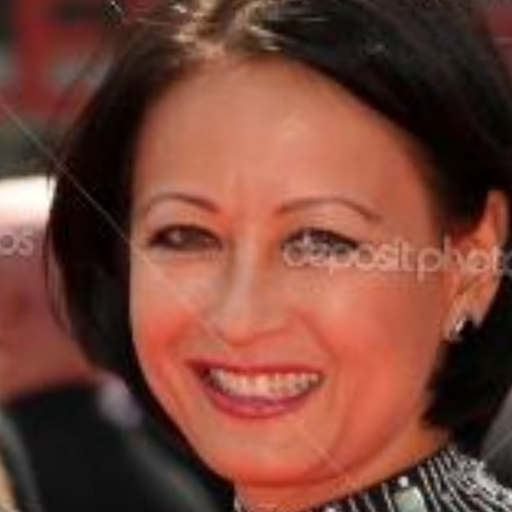}
\end{minipage}
\hspace{\sizeHorizontalSpace}
\begin{minipage}[b]{0.11\textwidth}\centering
    \includegraphics[width=\textwidth]{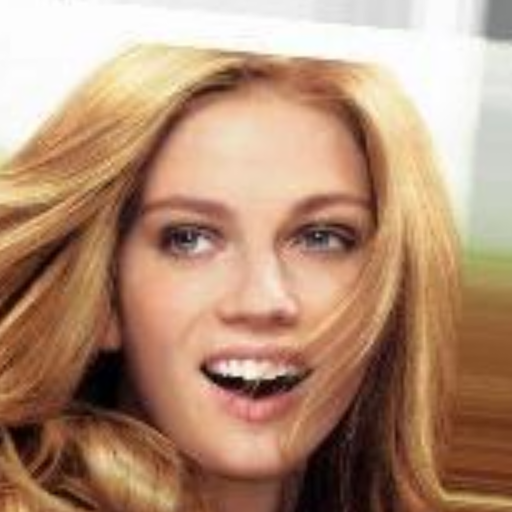}
\end{minipage}
\hspace{\sizeHorizontalSpace}
\begin{minipage}[b]{0.11\textwidth}\centering
    \includegraphics[width=\textwidth]{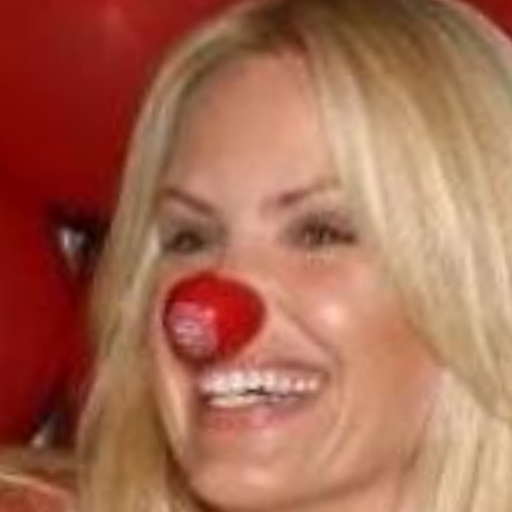}
\end{minipage}
\hspace{\sizeHorizontalSpace}
\begin{minipage}[b]{0.11\textwidth}\centering
    \includegraphics[width=\textwidth]{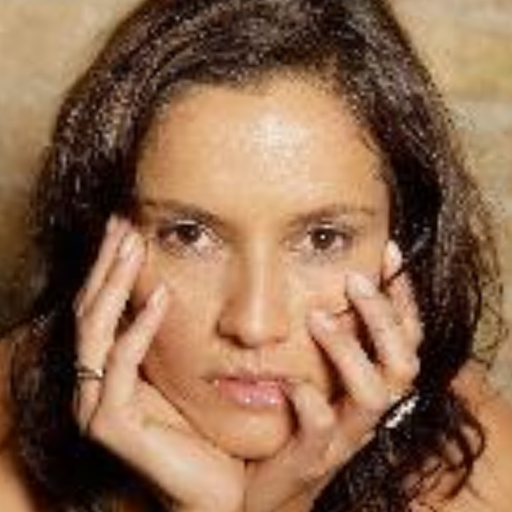}
\end{minipage}
\\
\begin{minipage}[b]{0.11\textwidth}\centering
    \includegraphics[width=\textwidth]{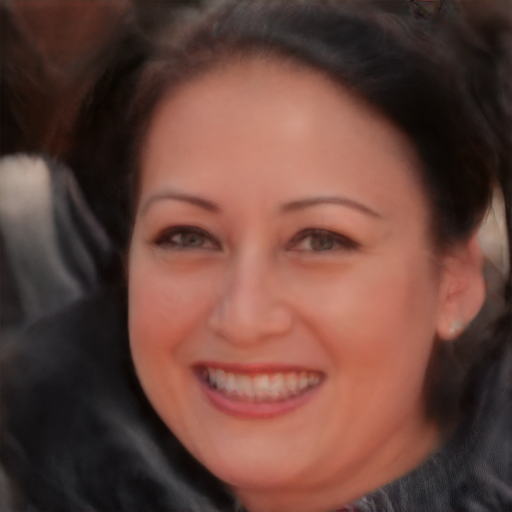}
\end{minipage}
\hspace{\sizeHorizontalSpace}
\begin{minipage}[b]{0.11\textwidth}\centering
    \includegraphics[width=\textwidth]{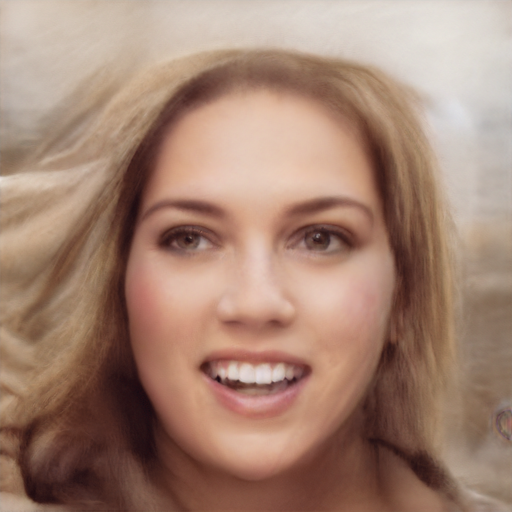}
\end{minipage}
\hspace{\sizeHorizontalSpace}
\begin{minipage}[b]{0.11\textwidth}\centering
    \includegraphics[width=\textwidth]{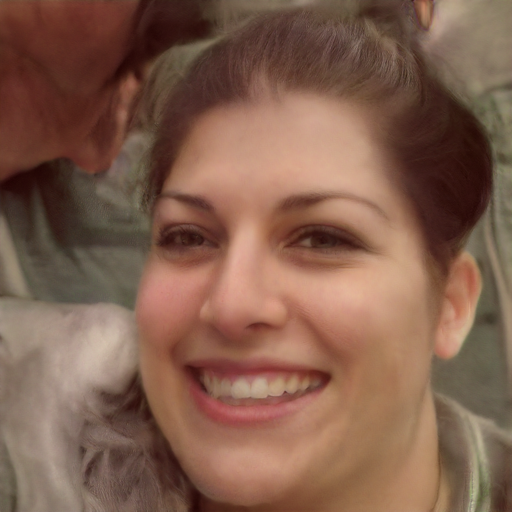}
\end{minipage}
\hspace{\sizeHorizontalSpace}
\begin{minipage}[b]{0.11\textwidth}\centering
    \includegraphics[width=\textwidth]{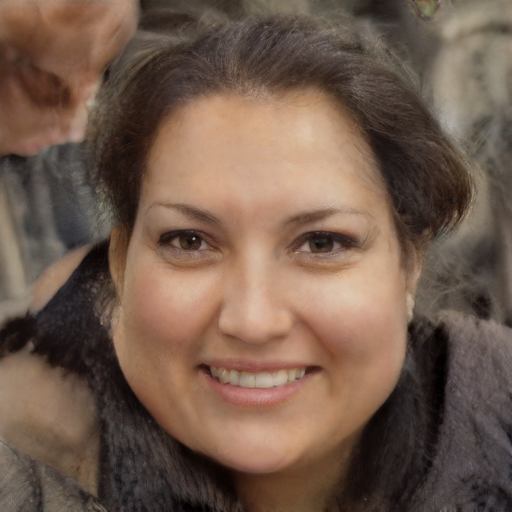}
\end{minipage}
\end{center}
\caption{
    Deep-blurring effects.
    By smoothing in latent space, DeepBlur removes artifacts and occlusions in the images at semantic level.
    For example, \textbf{top} are original images and \textbf{bottom} are the deep-blurred counterparts, where
    watermark, hair, nose sleeve, and hands are removed from the frontal faces, respectively.
}
\label{fig:deepblur_effects}
\end{figure}

\begin{figure}[t]
\begin{center}
\begin{minipage}[b]{0.11\textwidth}\centering
    \includegraphics[width=\textwidth]{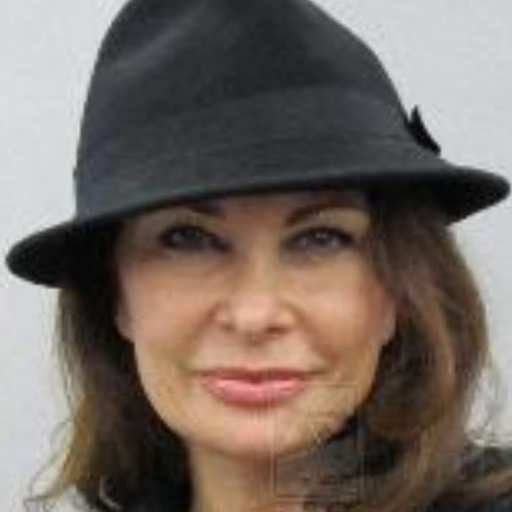}
\end{minipage}
\hspace{\sizeHorizontalSpace}
\begin{minipage}[b]{0.11\textwidth}\centering
    \includegraphics[width=\textwidth]{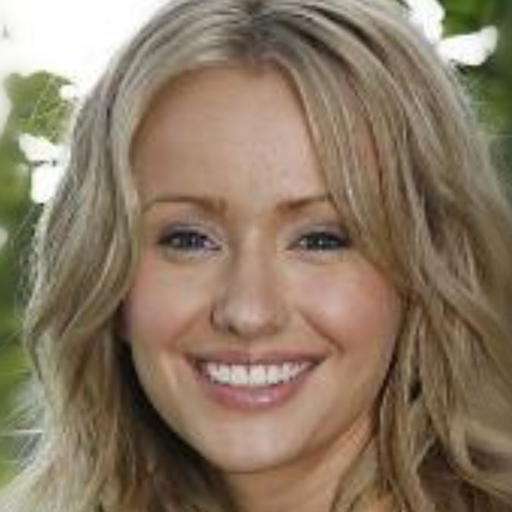}
\end{minipage}
\hspace{\sizeHorizontalSpace}
\begin{minipage}[b]{0.11\textwidth}\centering
    \includegraphics[width=\textwidth]{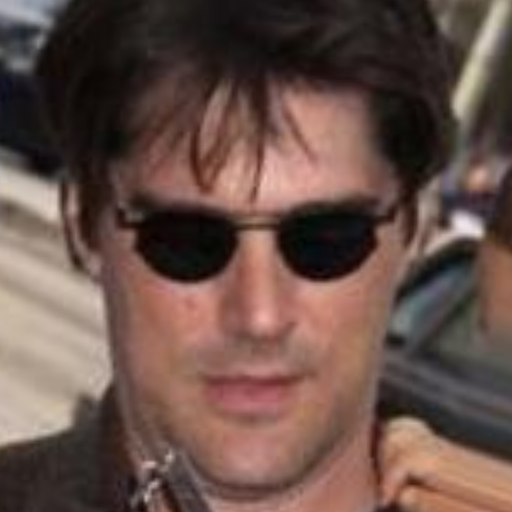}
\end{minipage}
\hspace{\sizeHorizontalSpace}
\begin{minipage}[b]{0.11\textwidth}\centering
    \includegraphics[width=\textwidth]{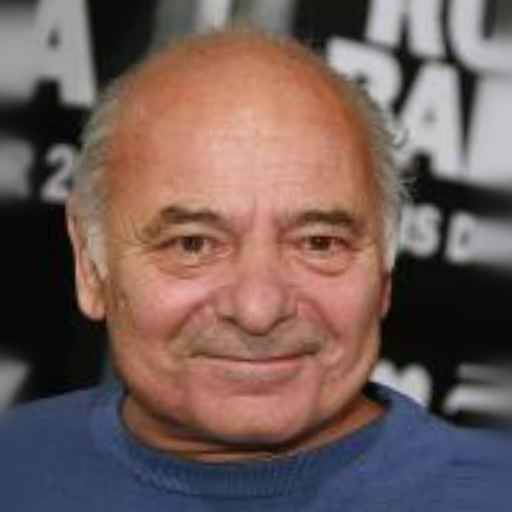}
\end{minipage}
\\
\begin{minipage}[b]{0.11\textwidth}\centering
    \includegraphics[width=\textwidth]{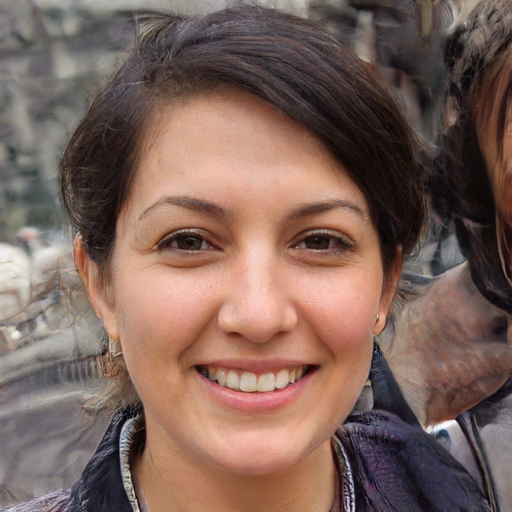}
\end{minipage}
\hspace{\sizeHorizontalSpace}
\begin{minipage}[b]{0.11\textwidth}\centering
    \includegraphics[width=\textwidth]{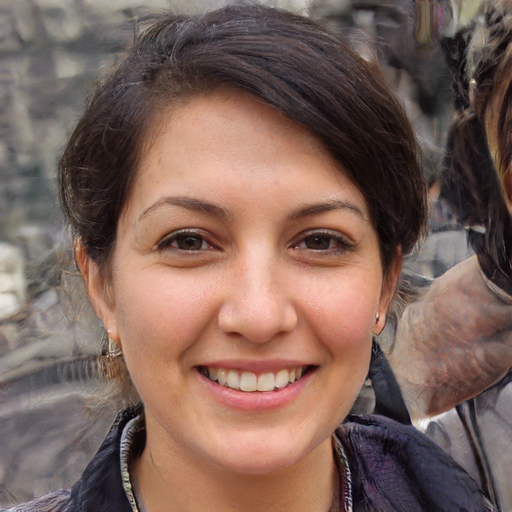}
\end{minipage}
\hspace{\sizeHorizontalSpace}
\begin{minipage}[b]{0.11\textwidth}\centering
    \includegraphics[width=\textwidth]{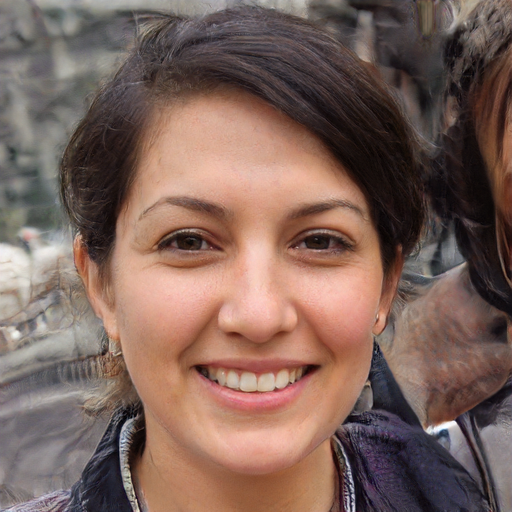}
\end{minipage}
\hspace{\sizeHorizontalSpace}
\begin{minipage}[b]{0.11\textwidth}\centering
    \includegraphics[width=\textwidth]{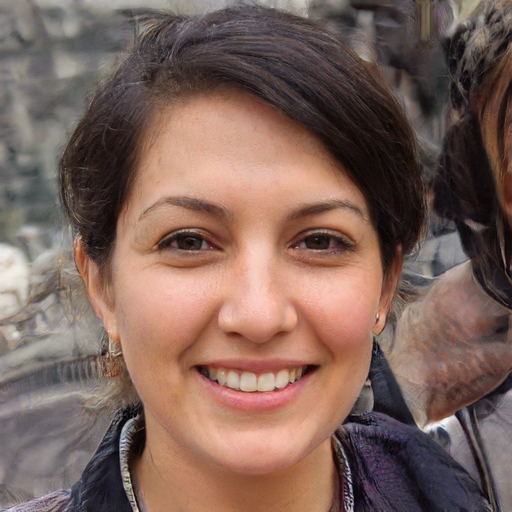}
\end{minipage}
\end{center}
\caption{
    Averaging effects.
    If we apply a very large kernel (e.g., $\sigma = 100$),
    the model will take the average of almost all latent values
    and generate an ``average face'' of GAN.
    In above examples,
    \textbf{top} are the original images and \textbf{bottom} are the averaged ones.
    Although the inputs are different, the averaged images are almost identical
    and only have subtle differences in background.
}
\label{fig:average_faces}
\end{figure}

first-order algorithms, e.g.,
stochastic gradient descent (SGD) \cite{QIAN1999145},
adaptive gradient algorithm (AdaGrad) \cite{duchi2011adaptive},
Adam \cite{kingma2014adam};
and second-order algorithms such as BFGS \cite{fletcher2013practical}.
To accelerate convergence, SGD adds momentum of previous weight when updating the current;
AdaGrad leverages adaptive learning rates;
Adam combines momentum with the adaptive learning rate method;
and BFGS approximates second-order derivatives and uses them for weight updates.

In general, second-order methods require more computing resources
per step as higher order information is required.
However, they usually take less steps to converge,
especially for functions close to convex, and robust against saddle points,
which is the case when we start the latent representation search from an ``average face.''
\Cref{fig:latent_code_search} shows that the search algorithm converges very quickly with an appropriate initialization,
and a latent representation obtained after only 10 steps can be used by the generator to synthesize
an image that looks close to the original.
\Cref{fig:optim_method_compare} compares the discussed methods in terms of numbers of iterations and elapsed time versus losses,
and demonstrates that L-BFGS (limited memory BFGS) has the best performance among the four, which aligns with our intuition
and thus it is used in our framework for all the experiments.

\subsection{Deep Blurring Effects}\label{sec:deepblur_effects}

In the experiments, we also observe some interesting visual effects by deep blurring,
which provide deeper insight of the method.
For example, \cref{fig:deepblur_effects} shows that deep blurring may remove artifacts and occlusions on frontal faces
when applying a low-pass filter to the latent representation that control the semantics.
\Cref{fig:average_faces} shows the ``average face'', which is achieved by filtering with a very large kernel size
(i.e., taking the average of all latent values).
We found that using the latent representation of the ``average face'' as initial value
for latent representation search (see \cref{fig:latent_code_search}), instead of random initialization,
makes the searching step converge faster, which can be explained by the property of second-order methods
we discussed in \cref{sec:complexity}.
These visual effects, along with results shown in \cref{fig:teaser,fig:method,fig:latent_code_search,fig:optim_loss,fig:deepblur_demo},
align with empirical findings in literature,
and provide new evidence of linearity and continuity in the latent space of GANs.

\section{Conclusion}\label{sec:conclusion}

To conclude the paper,
we present DeepBlur, a simple yet effective method for natural image obfuscation.
By blurring the latent space of a generative model,
DeepBlur is able to alter the identity in the image while preserving high visual quality.
We evaluate the method both qualitatively and quantitatively, and
show that it is effective against both human perception and state-of-the-art facial recognition systems.
Our experiments demonstrate that
DeepBlur has advantages in either image quality, computational efficiency,
effectiveness against unauthorized identification attacks, or all of the above
when comparing to established methods,
In the future, we plan to extend our method to broader applications.

{\small
\bibliographystyle{ieee_fullname}
\bibliography{db.bib}
}

\end{document}